\documentclass[lettersize,journal]{IEEEtran}
\usepackage{amsmath,amsfonts}
\usepackage{algorithmic}
\usepackage{algorithm}
\usepackage{hyperref}
\usepackage{array}
\usepackage[caption=false,font=normalsize,labelfont=sf,textfont=sf]{subfig}
\usepackage{textcomp}
\usepackage{stfloats}
\usepackage{multirow}
\usepackage{url}
\usepackage{verbatim}
\usepackage{graphicx}
\usepackage{subcaption}
\usepackage{orcidlink}
\usepackage[export]{adjustbox} 
\usepackage{cite}
\usepackage{todonotes}

\begin{document}

\title{ChromaFormer: A Scalable and Accurate Transformer Architecture for Land Cover Classification}

\author{
Mingshi Li$^{1}$~\orcidlink{0000-0002-4458-9051}, 
Dusan Grujicic$^{1}$~\orcidlink{0000-0002-7711-7528}, 
Ben Somers$^{2}$~\orcidlink{0000-0002-7875-107X}, 
Stien Heremans$^{3}$~\orcidlink{0000-0002-5356-1093}, 
Steven De Saeger$^{3}$~\orcidlink{0000-0002-5477-3063}, 
Matthew B. Blaschko$^{1}$~\orcidlink{0000-0002-2640-181X} 
\thanks{$^{1}$ESAT-PSI, KU Leuven, Belgium.}
\thanks{$^{2}$Department of Earth and Environmental Sciences, KU Leuven, Belgium.}
\thanks{$^{3}$Research Institute for Nature and Forest (INBO), Belgium.}}



\maketitle

\begin{abstract}
Remote sensing imagery from systems such as Sentinel provides full coverage of the Earth’s surface at around 10-meter resolution. The remote sensing community has transitioned to extensive use of deep learning models due to their high performance on benchmarks such as UCMerced and ISPRS Vaihingen datasets. Convolutional models such as UNet and ResNet variations are commonly employed for remote sensing but typically only accept three channels, as they were developed for RGB imagery, while satellite systems provide more than 10. Recently, several transformer architectures have been proposed for remote sensing, but they have not been extensively benchmarked and are typically used on small datasets such as Salinas Valley. Meanwhile, it is becoming feasible to obtain dense spatial land-use labels for entire first-level administrative divisions of some countries. Scaling law observations suggest that substantially larger, multi-spectral transformer models could provide a significant leap in remote sensing performance in these settings.

In this work, we propose ChromaFormer, a family of multi-spectral transformer models, which we evaluate across orders of magnitude differences in model parameters to assess their performance and scaling effectiveness on a densely labeled imagery dataset of Flanders, Belgium, covering more than 13,500 km$^2$ and containing 15 classes. We propose a novel multi-spectral attention strategy and demonstrate its effectiveness through ablations. Furthermore, we show that models many orders of magnitude larger than conventional architectures, such as UNet, lead to substantial accuracy improvements: a UNet++ model with 23M parameters achieves less than 65\% accuracy, while a multi-spectral transformer with 655M parameters achieves over 95\% accuracy on the Biological Valuation Map of Flanders.

\end{abstract}


\section{Introduction}
\IEEEPARstart{R}{emote} sensing plays a crucial role in environmental monitoring, urban planning, disaster forecasting, and more by utilizing rich data from satellite and aerial systems. Processing this vast, high-dimensional data poses significant challenges, especially with traditional, time-consuming, and error-prone techniques. Recent advances in machine learning (ML), particularly convolutional neural networks (CNNs), have greatly enhanced the accuracy and efficiency of remote sensing analysis by automatically learning complex spatial and spectral patterns \cite{zhang2016deep, li2019deep}.

CNNs, originally developed for image tasks like classification, detection, and segmentation, have been effectively introduced into remote sensing. For instance, \cite{huhyper} demonstrated the effectiveness of CNNs in hyperspectral image classification, achieving state-of-the-art performance at that time. CNNs have since been applied to land-use classification, terrain change detection, and urban planning.

Transformers, initially successful in natural language processing \cite{vaswani2017attention}, have been adapted for vision tasks, improving the handling of distant dependencies and complex spatial correlations via attention modules \cite{carion2020end, han2022survey}. \cite{dosovitskiy2020image} introduced the Vision Transformer (ViT), which outperforms CNN-based models on various image classification benchmarks, suggesting its potential for remote sensing tasks. \cite{liu2021swin} designed shifted window transformers (Swin-T), providing an energy-efficient variation of transformer. \cite{stunet} later demonstrated their superiority over conventional CNNs in multi-modal remote sensing data. These works highlight the importance of architectural choices in neural networks for remote sensing.

Despite these advancements, the scaling properties of neural networks in remote sensing remain under-explored. While larger models and datasets might intuitively lead to better performance, the relationship between model size, data size, and task performance in remote sensing contexts has not been thoroughly investigated. Exploring these scaling laws \cite{DBLP:journals/corr/abs-2001-08361} is necessary to develop efficient and accurate models.

In this paper, we first introduce the current state of research in remote sensing machine learning and point out the model and dataset mismatching issues that mainstream machine learning applications currently face. Then we propose a spatial-spectral module fused Swin Transformer backbone for multi-spectral segmentation tasks. We systematically explore the scaling laws in remote sensing and demonstrate that our backbone achieves state-of-the-art performance by conducting an extensive empirical study on a selection of neural network architectures. Furthermore, we investigate how performance varies with model size, architecture selection, and training data size. Additionally, we perform an ablation study to examine the impact of different components on the scaling behaviors of exemplary simple architectures. Our results provide meaningful insights on how to scale neural network models according to specific application needs in multi-spectral remote sensing tasks.

\section{Background and motivation}
\subsection{Machine Learning in Remote Sensing}
Machine learning (ML) has significantly changed remote sensing research with deep learning methods, and convolutional neural networks (CNNs) are growing to be central to processing and analyzing complex remote sensing data. Early applications of ML in remote sensing focused on using traditional algorithms, such as support vector machines and decision trees, for simple tasks like land cover classification and change detection. As more accurate and powerful tools and architecture design choices came out in recent years, the implementation of CNNs has led to substantial performance improvements in various complicated tasks such as image classification, object detection, and semantic segmentation \cite{zhureview}. CNNs also showed effectiveness in processing hyperspectral images and significantly outperformed traditional methods by learning hierarchical feature representations directly from data \cite{huhyper}. For high-resolution satellite imagery, \cite{maggiori2016convolutional} has proven that tailor-made deep CNNs can handle large-scale remote sensing image classification and achieve state-of-the-art accuracies.

Transformer models have further reshaped and accelerated the advances in remote sensing research with even better accuracy and abilities to handle complex input samples. Originally designed for natural language processing tasks, transformers were later adapted to vision tasks as Vision Transformers (ViT) by \cite{dosovitskiy2020image}, outperforming CNNs in some tasks by leveraging long-range spatial dependencies with attention mechanisms. Because of its superior performance on image classification tasks, ViT showed potential applicability to remote sensing tasks. Their work was later extended by applying transformers to multi-modal remote sensing data, demonstrating architectural design choices are crucial for model performance across different data modalities and scales \cite{transsurvey}.

Enlightened by the spatial attention mechanism in Transformer architectures, researchers have been trying to adapt this mechanism for multi/hyper-spectral remote sensing tasks actively. \cite{hangssatt} introduced spectral-attention-aided CNN models and proved that it is beneficial to include spectral-attention modules into CNN backbones. Later work of \cite{resssatt} demonstrated that ResNet can be improved by incorporating spectral attention modules. Moreover, \cite{fas} designed a spectral association block that focuses on establishing the connection between different locations in the cuboid by calculating spectral association kernels using 3D convolutions. Although spectral information is included in the training process, their methods are all specified in sparse representations of spectral correlation but not learning from token-based to spectral band-based inter-channel attention.

\subsection{Dataset size dominates model selection}

\begin{figure*}[htbp]
\begin{center}
    \includegraphics[width=7in]{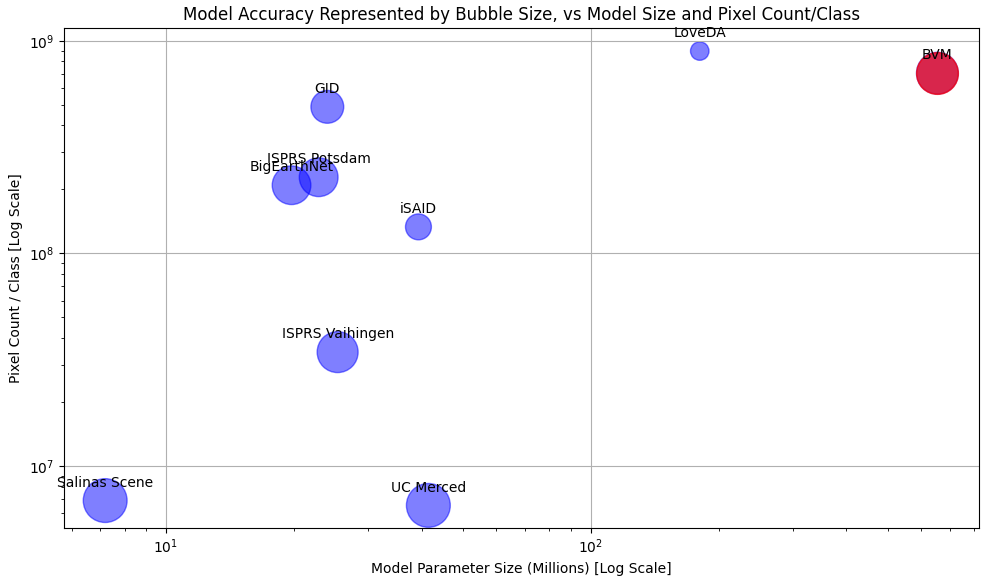}
\end{center}
\caption{Comparison of the size of mainstream remote sensing datasets and the BVM dataset \cite{bvm} (dataset size vs. model size). Y-axis represents the overall dataset size in the form of total pixel number per class in each dataset. X-axis is the averaged parameter size of models used on each of the datasets. The size of bubbles represents averaged accuracy scores. The BVM dataset is placed at the top-right corner, indicating both the model size and dataset size are well above the current norms of the latest research.} 
\label{fig:ourscale}
\end{figure*}

Remote sensing and machine learning researchers seem to have been stranded by the ideas of developing novel adaptations of existing deep learning architectures and sophisticated numerical methods to improve the performance of individual models on individual datasets \cite{bergamasco2023dual, peng2023rsbnet, roy2023multimodal,yuan2023litest,lv2023shapeformer,zhang2023efficient}. Few have studied the scaling properties of models that are used, especially in the remote sensing field where more and more very large datasets are emerging. It is important to point out that in the current state of remote sensing research, most models being employed are relatively small, typically containing fewer than 200 million parameters. Taking ResNet50, for example, with a small size of 25 million parameters, it has been actively applied to datasets ranging from tiny ones like Salinas Scenes \cite{salinas}, to very large datasets such as BigEarthNet \cite{clasen2024reben}. Even after the adaptation of large vision transformers to remote sensing, these models are still applied to datasets of varying scales without considering the efficiency and performance gains. It became necessary to do empirical studies on a very large dataset with spatial-spectral models of different sizes to advocate a paradigm that matching the scaling properties of models with dataset size is crucial for benchmarking models properly.

\begin{table*}[htbp]
\centering
\caption{A comparison of mainstream datasets and BVM dataset, and the models applied to them. Note that the pixel count is an estimation based on dataset specifications, and not all datasets and models are covered.}
\setlength{\tabcolsep}{4pt}  
\begin{tabular}{|c|c|c|c|c|c|}
\hline
\textbf{Dataset} & \textbf{Pixel Count} & \textbf{Classes} & \textbf{Model} & \textbf{Accuracy} & \textbf{Reference} \\ \hline

\multirow{4}{*}{Salinas Scene} 
& \multirow{4}{*}{0.11M} & \multirow{4}{*}{16} 
& PSE-UNet & 91.01\% (OA) & \cite{li2022}  \\ \cline{4-6}
& & & 3D-CNN & 97.55\% (OA) & \cite{3dcnn} \\ \cline{4-6}
& & & HybridSN & 99.84\% (OA) & \cite{hybridsn}  \\ \cline{4-6}
& & & SMALE & 99.28\% (OA) & \cite{smale} \\ \hline

\multirow{4}{*}{UC Merced} 
& \multirow{4}{*}{137M} & \multirow{4}{*}{21} 
& DenseNet-121 & 99.88\% (OA) & \cite{agos}  \\ \cline{4-6}
& & & MS2AP & 99.01\% (OA) & \cite{ms2ap} \\ \cline{4-6}
& & & LSENet & 98.69\% (OA) & \cite{lsenet}  \\ \cline{4-6}
& & & VGG-VD16 & 95.21\% (OA) & \cite{vggvd16} \\ \hline

\multirow{4}{*}{ISPRS Vaihingen} 
& \multirow{4}{*}{206M} & \multirow{4}{*}{6} 
& PGNet & 86.32\% (OA) & \cite{pgnet}  \\ \cline{4-6}
& & & MANet & 86.51\% (OA) & \cite{manet} \\ \cline{4-6}
& & & EMNet & 95.42\% (OA) & \cite{emnet}  \\ \cline{4-6}
& & & DeepLabv3+ & 86.07\% (OA) & \cite{deeplabv3} \\ \hline

\multirow{5}{*}{ISPRS Potsdam} 
& \multirow{5}{*}{1.37B} & \multirow{5}{*}{6} 
& CM-UNet & 91.86\% (OA) & \cite{cmunet}  \\ \cline{4-6}
& & & SSCNet & 91.03\% (OA) & \cite{sscnet} \\ \cline{4-6}
& & & HCANet & 90.15\% (OA) & \cite{hcanet}  \\ \cline{4-6}
& & & AerialFormer-B & 91.4\% (OA) & \cite{rs16162930} \\ \cline{4-6}
& & & DC-Swin & 92\% (OA) & \cite{dcswin} \\ \hline

\multirow{3}{*}{iSAID} 
& \multirow{3}{*}{2B} & \multirow{3}{*}{15} 
& SegNeXt-L & 70.3\% (IoU) & \cite{segnext}  \\ \cline{4-6}
& & & SegNeXt-B & 69.9\% (IoU) & \cite{segnext} \\ \cline{4-6}
& & & AerialFormer-B & 69.3\% (IoU) & \cite{rs16162930} \\ \hline

\multirow{3}{*}{LoveDA} 
& \multirow{3}{*}{6.27B} & \multirow{3}{*}{7} 
& UNet-Ensemble & 56.16\% (IoU) & \cite{unetensemble}  \\ \cline{4-6}
& & & SFA-Net & 54.9\% (IoU) & \cite{sfanet} \\ \cline{4-6}
& & & ViT-G12X4 & 54.4\% (IoU) & \cite{vitg12x4} \\ \hline

\multirow{3}{*}{GID} 
& \multirow{3}{*}{7.34B} & \multirow{3}{*}{15} 
& LSKNet-S & 82.3\% (OA) & \cite{lsknet}  \\ \cline{4-6}
& & & DeepTriNet & 77\% (OA) & \cite{deeptrinet}  \\ \hline

\multirow{3}{*}{BigEarthNet} 
& \multirow{3}{*}{9B} & \multirow{3}{*}{43} 
& ResNet50 & 91.8\% (OA) & \cite{ssleo}  \\ \cline{4-6}
& & & ViT-S & 89.9\% (OA) & \cite{ssleo} \\ \cline{4-6}
& & & ResNet18 & 89.3\% (OA) & \cite{selfs}  \\ \hline

\multirow{1}{*}{\textbf{BVM}} 
& \multirow{1}{*}{\textbf{10.57B}} & \multirow{1}{*}{\textbf{15}} 
& \textbf{Swin-h} & \textbf{96.71\% (OA)} & \cite{bvm} \\ \hline

\end{tabular}

\label{tab:scale_comp}
\end{table*}

In this paper, we use the Biological Valuation Map (BVM) of Flanders s\cite{bvm, 1df9a7c2964043d5b80a06fcde793bbd}, a recently published land cover dataset with pixel-wise labeling, in combination with Sentinel-2 public imagery provided by ESA, as the training dataset. The scale of the experiment positions our work in a different region of the research field, as seen in \autoref{fig:ourscale} and Appendix A.1, where larger models and larger datasets are needed to push the boundaries of performance instead of sticking to small datasets and small model studies. For example, datasets such as UCMerced (with 137M pixels) and ISPRS Vaihingen (with around 169M pixels) are widely used for benchmarking models of which parameter size could range from but are considerably smaller than BVM dataset in both pixel count and complexity. Similarly, the ISPRS Potsdam dataset contains roughly 1.37 billion pixels, which, while larger than Vaihingen, is still much smaller than the scale we are dealing with (\autoref{tab:scale_comp}). There are also larger satellite datasets like LoveDA and BigEarthNet, but their labels are either unreliable or sparse. Our research operates on a different scale, focusing on advancing large-scale segmentation tasks.

\begin{figure*}[h]
  \centering
  \includegraphics[width=0.8\textwidth]{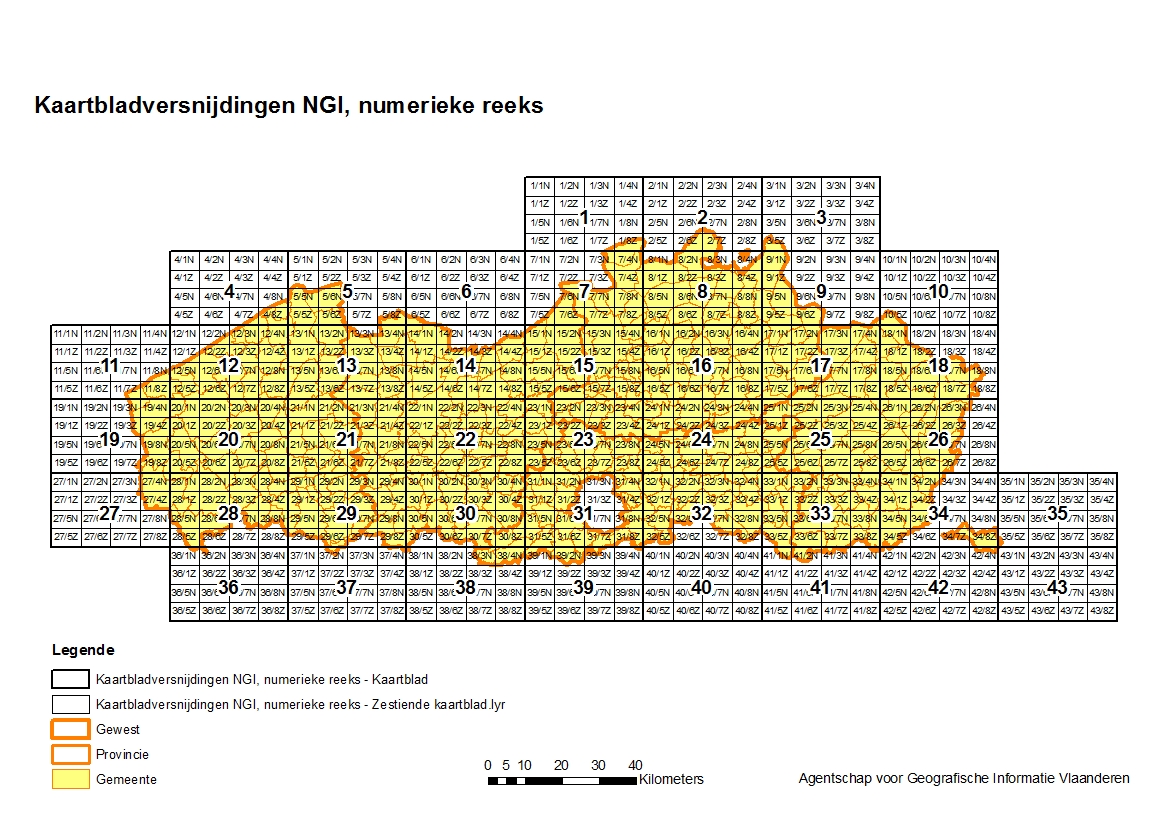}
  \caption{A demonstration of map sheet cutouts (Source: Digital Flanders Agency): the Flanders region is confined in 43 blocks, each block is further partitioned into 16 smaller ones where 8 on the left is marked for training, 4 on the upper right marked for validation and 4 on the lower right marked for test}
  \label{fig:kaart}
\end{figure*}

In terms of preparation of the BVM data, we introduce the topographic map sheet cutouts, the "kaartbladenversnijdingen (map sheet divisions)" (\autoref{fig:kaart}), to help us partition the BVM map further for training, validation, and test dataset splits. As the following figure shows: for each number X indexed block, we group X/1N, X/2N, X/1Z, X/2Z, X/5N, X/6N, X/5Z, X/6Z together as training blocks; X/13N, X/4N, X/3Z, X/4Z as validation blocks and X/17N, X/8N, X/7Z, X/8Z as test blocks. Each training or validation/test block covers a land area of size 160 $km^2$.

Using a kaartbladen grid to split up Flanders and choosing corresponding blocks is a robust way to partition training, validation, and test datasets. To ensure homogeneity in the class distribution of these three datasets, we report the Chi-squared distance between them. Chi-squared distance is often used to determine the similarity of two categorical distributions; it is formulated as follows:
$D_{\chi^2}(P, Q) = \sum_{i} \frac{(P(i) - Q(i))^2}{P(i) + Q(i)}$,
where P(i) and Q(i) are the probabilities of the i-th element in distributions P and Q, respectively.

The number of pixels for each class in three datasets and the total number of pixels are taken. The Chi-squared distance can thus be calculated as 0.0038 between training and validation sets, 0.00392 between training and test sets, and 0.0048 between validation and test sets. All three values are relatively low, indicating that the class distributions of all three sets are very similar to each other. Such close distances are desirable in our case as it suggests that the validation and test sets are good representatives of the training data. Details for class distribution are in \autoref{tab:class_distribution}.

\begin{table}[h]
\centering
\small 
\begin{tabular}{|p{3cm}|r|r|r|}
\hline
\textbf{Class} & \textbf{Train (\%)} & \textbf{Val. (\%)} & \textbf{Test (\%)} \\ \hline
Coastal dune habitats & 0.10 & 0.06 & 0.25 \\ \hline
Cultivated land & 34.27 & 33.92 & 32.85 \\ \hline
Grasslands & 23.07 & 22.92 & 22.14 \\ \hline
Heathland & 0.54 & 1.08 & 0.95 \\ \hline
Inland marshes & 0.24 & 0.23 & 0.22 \\ \hline
Marine habitats & 0.26 & 0.04 & 0.34 \\ \hline
Pioneer vegetation & 0.69 & 0.65 & 0.56 \\ \hline
Small landscape features - not specified & 0.05 & 0.06 & 0.07 \\ \hline
Small non-woody landscape features & 0.14 & 0.13 & 0.12 \\ \hline
Small woody landscape features & 0.63 & 0.58 & 0.73 \\ \hline
Unknown & 0.01 & 0.007 & 0.006 \\ \hline
Urban areas & 26.27 & 27.05 & 28.56 \\ \hline
Water bodies & 2.08 & 2.05 & 1.74 \\ \hline
Woodland and shrub & 11.65 & 11.22 & 11.47 \\ \hline
\end{tabular}
\caption{Class distribution ratios in train, validation, and test Datasets}
\label{tab:class_distribution}
\end{table}

\subsection{Scaling architectures on multi-spectral remote sensing tasks}

Scaling laws demonstrate that increasing model size, data volume, and computing improves neural network performance, as first formalized by \cite{DBLP:journals/corr/abs-2001-08361} and extended to generative models \cite{henighan2020scaling} and transfer learning \cite{hernandez2021scaling}. The CNN model architecture can greatly affect its effectiveness in remote sensing tasks; determining hyperparameters such as the depth, width, and input dimensions can often be up to the choices of engineers. EfficientNet introduced by \cite{tan2019efficientnet} pioneered architectural scaling by introducing a compound scaling method with a fixed set of hyperparameter coefficients. This approach yielded state-of-the-art results on multiple benchmarks with fewer parameters and reduced computational cost.

Meanwhile, \cite{rosenfeld2021predictability} examines how the performance of pruned neural networks scales with model size, data volume, and compute resources, highlighting the importance of matching model complexity with data to prevent overfitting and inefficient computation. \cite{zhang2022resnest} further explored the relationship between architectural design and scaling laws, confirming that deeper and wider models tend to benefit more than shallow and narrow ones, suggesting the importance of architectural design in optimizing model performances. Although the scaling of vision transformers has been studied \cite{zhai2022scaling}, the problem in multi-spectral vision tasks could be influenced by the composition of different modules in models. Our study fills the gap between their holistic analysis and multi-spectral cases.

\section{Methodologies}
\label{headings}

In this section, we first introduce the key component of ChromaFormer: the spectral dependency module (SDM). Then we show our SDM-infused scaling-efficient Swin transformer backbone that was adapted for large-scale multi-spectral training. We define a scaling factor that provides a quantitative approach to measure the scalability of architectures.

\subsection{Spectral dependency module}
We introduce a Spectral Dependency Module (SDM), whose design is similar to the spatial attention head in the Transformer, modifying the traditional attention mechanism by replacing token-based attention with spectral band-based attention. Instead of treating tokens as the basic unit of input, the SDM module considers spectral bands in multi-spectral data as the fundamental elements. 
\begin{figure}[htbp]
\begin{center}
    \includegraphics[width=3in]{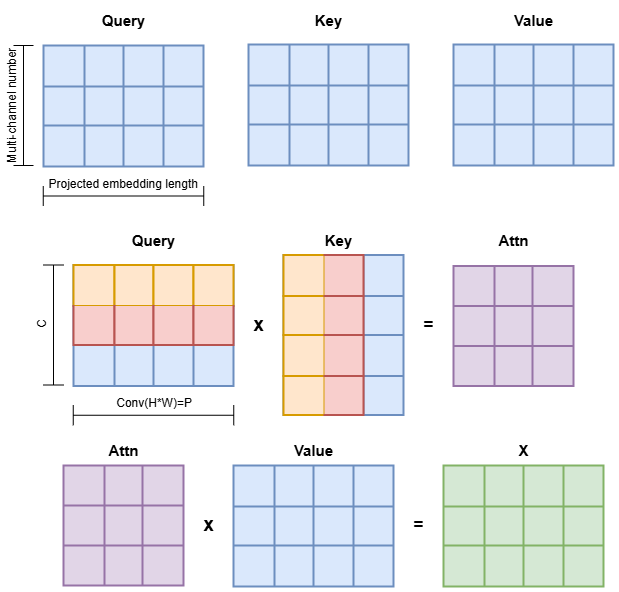}
\end{center}
\caption{SDM block mechanism, similar to transformer attention head, the query token number dimension is substituted by multi-spectral channel number. Embeddings from same channels are marked in the same color.} 
\label{fig:spec_corr}
\end{figure}
Each spectral band is rearranged into patches, and these patches of each channel are projected into an embedding space, similar to how tokens are projected in classical attention mechanisms. The core of this module is to compute the channel dependencies between all spectral bands across different channels. By computing attention weights based on spectral dependencies, the SDM allows for a fine-grained representation of spectral relationships, enabling improved performance in tasks requiring multi-spectral data processing.

Let \( X \in \mathbb{R}^{C \times H \times W} \) denote the multi-spectral data, where:
\begin{itemize}
    \item \( C \) is the number of spectral bands (channels),
    \item \( H \) and \( W \) are the spatial dimensions (height and width).
\end{itemize}

For each spectral band \( c \in \{1, 2, \dots, C\} \), we rearrange its data \( X_c \in \mathbb{R}^{H \times W} \) into a set of patches. The patches are represented as:
\[
\mathbf{x}_{c,p} \in \mathbb{R}^{P}, \quad p \in \{1, 2, \dots, N_p\}.
\]

Each patch \( \mathbf{x}_{c,p} \) is projected into an embedding space using a linear projection function \( E: \mathbb{R}^{P} \rightarrow \mathbb{R}^{d} \):
\[
\mathbf{z}_{c,p} = E(\mathbf{x}_{c,p}) \in \mathbb{R}^{d}.
\]

For each patch index \( p \), we stack the embeddings of all spectral bands to form a matrix:
\[
\mathbf{Z}_p = \begin{bmatrix}
\mathbf{z}_{1,p} \\
\mathbf{z}_{2,p} \\
\vdots \\
\mathbf{z}_{C,p}
\end{bmatrix} \in \mathbb{R}^{C \times d}.
\]

We compute the query (\( \mathbf{Q}_p \)), key (\( \mathbf{K}_p \)), and value (\( \mathbf{V}_p \)) matrices for each patch \( p \) using learnable weight matrices \( \mathbf{W}_Q \), \( \mathbf{W}_K \), and \( \mathbf{W}_V \):
\[
\begin{aligned}
\mathbf{Q}_p &= \mathbf{Z}_p \mathbf{W}_Q \in \mathbb{R}^{C \times d_k}, \\
\mathbf{K}_p &= \mathbf{Z}_p \mathbf{W}_K \in \mathbb{R}^{C \times d_k}, \\
\mathbf{V}_p &= \mathbf{Z}_p \mathbf{W}_V \in \mathbb{R}^{C \times d_v}.
\end{aligned}
\]

The correlation weights are calculated by computing the scaled dot-product attention over the spectral bands for each patch:
\[
\mathbf{A}_p = \text{softmax}\left( \frac{\mathbf{Q}_p \mathbf{K}_p^\top}{\sqrt{d_k}} \right) \in \mathbb{R}^{C \times C}.
\]
The softmax function is applied row-wise to ensure that the attention weights for each spectral band sum to 1. By using the attention weights \( \mathbf{A}_p \), we compute the output embeddings for each patch:
\[
\mathbf{O}_p = \mathbf{A}_p \mathbf{V}_p \in \mathbb{R}^{C \times d_v}.
\]

The outputs \( \mathbf{O}_p \) from all patches are aggregated by averaging to form the final representation used for downstream tasks:
\[
\mathbf{O} = \text{Avg}\left( \{\mathbf{O}_p\}_{p=1}^{N_p} \right).
\]
\begin{figure}[htbp]
\begin{center}
    \includegraphics[width=1.5in]{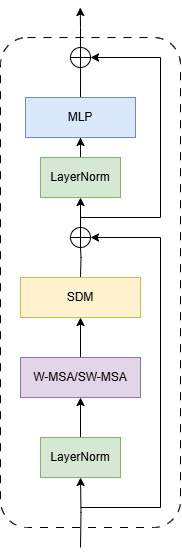}
\end{center}
\caption{The transformer block of a ChromaFormer model}
\label{fig:spatialspectral}
\end{figure}

SDM combined with classical vision transformer attention blocks can capture both spatial and inter-channel dependencies, thus enhancing spectral-spatial feature learning. It also introduces a scheme for building and scaling hybrid models by combining SDM blocks with classical attention heads and convolutional layers. This modular approach enables flexible model scaling by adjusting the number and configuration of SDM, attention, and convolution blocks.

\subsection{Architectures and Multi-Spectral Swin Transformer}

In this paper, we selected UNet++ \cite{zhou2018unet++}, ResNet \cite{he2016deep}, and Swin Transformer \cite{liu2021swin} as comparisons to ChromaFormer for scaling behavioral studies due to their wide usage in remote sensing tasks. These models also scale easily—ResNet by adding residual blocks, Transformers through layers or embedding dimensions, and our model via its modular design—allowing for detailed studies of scaling laws. Moreover, each model demonstrates consistent performance improvements with increasing scale, as seen with deeper ResNets \cite{he2016deep}, larger transformers \cite{NEURIPS2020_1457c0d6} and our experiments.

In this paper, we modified the Swin transformer models by connecting SDM to the multi-head self-attention and shift-window multi-head self-attention modules in each transformer block at stage 1. As \autoref{fig:spatialspectral} shows, the output dimension of MSA modules is the same as SDM's input dimensions, so it allows seamless connection of two modules. Our choice to insert SDM at stage 1 is due to the consideration that the first stage preserves the most raw spectral features that the SDM module may need. We refer to this modified architecture as ChromaFormer.

\subsection{Experiment specifications}

In this paper, instead of using an end-to-end training strategy, we choose to train the model with griding and patching \cite{chen2014deep} by first griding the large training images into smaller cells, then randomly sampling a certain number of cells to form up patches. This approach not only makes it feasible to train on high-resolution images by lowering memory requirements but also increases the diversity of training data, as multiple patches are extracted from a single image. By focusing on local patterns within these patches or grids, the strategy enhances the model's ability to learn fine-grained spatial features and improves overall generalization performance \cite{li2016hyperspectral}. We use Overall Accuracy (OA) as our metric for model performance because it provides a straightforward, global measure of how well the model correctly classifies pixels across all classes.

The scaling efficiency coefficient \( S \) quantifies how effectively a neural network scales its performance relative to the increase in parameters and computational resources. It is mathematically defined as:
\[
S = -\frac{1}{\log\left(\frac{G}{P \times C}\right)}
\]
where:
\begin{itemize}
    \item \( G \) is the Performance Gain Factor.
    \item \( P \) is the Parameter Count Scaling Factor.
    \item \( C \) is the Computation Increase Factor.
\end{itemize}

The proposed coefficient provides a quantitative measure of scaling efficiency for neural networks. By considering both the performance gain and the increase in the consumption of computational resources (parameter count and computation increase), the coefficient evaluates how effectively a model scales. A higher coefficient indicates that a model achieves a larger performance improvement with relatively modest increases in parameters and training time, making it more efficient in scaling. Conversely, a lower coefficient suggests that either the scaling process is inefficient or the model is approaching the limit of diminishing returns. At this stage scaling the model further with significant resource investments results in only marginal performance gains.

\section{Results and Discussion}
For all our experiments, we utilized a single-node setup equipped with four NVIDIA A100-80G GPUs. We adopted distributed data parallelism with shared gradient synchronization across the GPUs, and the gradient accumulation technique was managed through the HuggingFace Accelerate API, which allowed efficient distribution of the training load while reducing the memory requirements by accumulating gradients over multiple mini-batches before updating the model weights. 

For the optimizer, we employed the Adam optimizer because of its adaptive learning rate properties, making it well-suited for large-scale training tasks. All GPUs were contained within a single node to make sure training processes introduced no inter-node communications, and all input samples were treated in read-only mode, minimizing the possible I/O overhead. We reserve sufficient memory space regardless of the model size to keep our training settings constant and stable for the ablation study.

\begin{table*}[htbp]
\centering
\caption{Comparison of models with parameters, training time, accuracy (with 95\% error bars computed using \(N=21\,500\,000\) samples), and scaling coefficients.}
\begin{tabular}{|l|c|c|c|c|}
\hline
\textbf{Model} & \textbf{Parameters (M)} & \textbf{Time/Epoch (h)} & \textbf{Accuracy (\%)} & \textbf{Scaling Coefficient} \\ \hline
\multicolumn{5}{|c|}{\textbf{Small Models (~1M to 30M Parameters)}} \\ \hline
ResNet-1M        & 1   & 3.5  & 75.92 ± 0.02  & Baseline \\ \hline
ResNet-2M        & 2   & 3.9  & 76.03 ± 0.02  & 2.879    \\ \hline
UNet++           & 23  & 4.0  & 64.48 ± 0.02  & N/A      \\ \hline
ResNet-20M       & 20  & 4.1  & 80.95 ± 0.02  & 0.745    \\ \hline
Swint            & 27  & 8.7  & 91.34 ± 0.01  & Baseline \\ \hline
ChromaFormer-t   & 27  & 8.7  & 92.25 ± 0.01  & Baseline \\ \hline
\multicolumn{5}{|c|}{\textbf{Medium Models (~50M to 100M Parameters)}} \\ \hline
ResNet-230M      & 230 & 7.4  & 84.10 ± 0.02  & 0.378    \\ \hline
Swins            & 49  & 12.6 & 92.19 ± 0.01  & 2.406    \\ \hline
ChromaFormer-s   & 49  & 12.6 & 92.53 ± 0.01  & 2.390    \\ \hline
Swinb            & 86  & 14.8 & 93.08 ± 0.01  & 1.378    \\ \hline
ChromaFormer-b   & 86  & 14.8 & 93.38 ± 0.01  & 1.373    \\ \hline
\multicolumn{5}{|c|}{\textbf{Large Models (~150M to 300M Parameters)}} \\ \hline
ResNet-1550M     & 1550& 25.2 & 87.32 ± 0.02  & 0.251    \\ \hline
Swinl            & 195 & 16.3 & 94.57 ± 0.01  & 0.896    \\ \hline
ChromaFormer-l   & 195 & 16.3 & 94.80 ± 0.01  & 0.893    \\ \hline
\multicolumn{5}{|c|}{\textbf{Extra-Large Models (~650M to 2800M Parameters)}} \\ \hline
ResNet-2800M     & 2800& 40   & 89.19 ± 0.01  & 0.225    \\ \hline
Swinh            & 655 & 24.0 & 96.64 ± 0.01  & 0.555    \\ \hline
ChromaFormer-h   & 656 & 24.0 & 96.71 ± 0.01  & 0.554    \\ \hline
\end{tabular}
\label{tab:coeffs}
\end{table*}

\begin{figure*}[htbp]
\begin{center}
    \includegraphics[height=4in,width=6in]{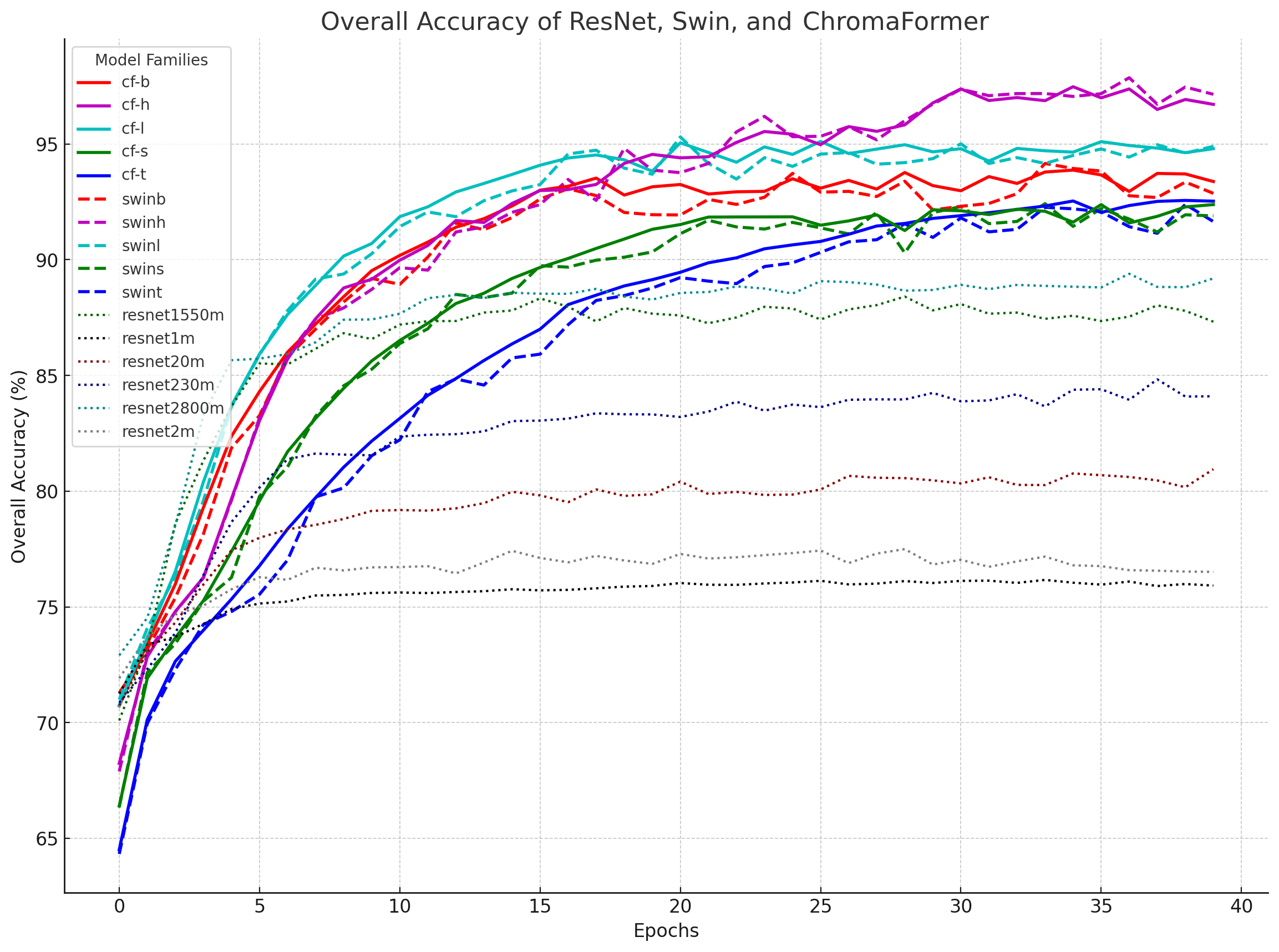}
\end{center}
\caption{Validation accuracy curves of scaled networks, "m" in legend stands for million parameters, and "epochs" stands for how many times the entire dataset has passed through the model.}
\label{fig:acc_curves}
\end{figure*}

As evident by \autoref{fig:acc_curves} and \autoref{tab:coeffs}, the Swin and ChromaFormer models exhibit superior performance compared to both the ResNet family and the UNet++ model across all sizes, based on the provided data. In the small model category, Swint (27 million parameters) achieves an accuracy of 91.34\%, significantly outperforming ResNet-20M (20 million parameters) with 80.95\% accuracy and UNet++ (23 million parameters) with 64.48\% accuracy. This substantial accuracy gap highlights the efficiency of Swin models in handling complex tasks with fewer parameters. Additionally, Swin models maintain higher scaling coefficients than ResNet models, indicating more efficient scaling as model size increases. The hierarchical architecture and shifted window mechanism of Swin transformers contribute to their enhanced performance and scalability, making them favorable choices over traditional convolutional models like ResNet and UNet++.

\begin{figure}[h]
\centering
\begin{tabular}{cc}
    \includegraphics[width=1.6in]{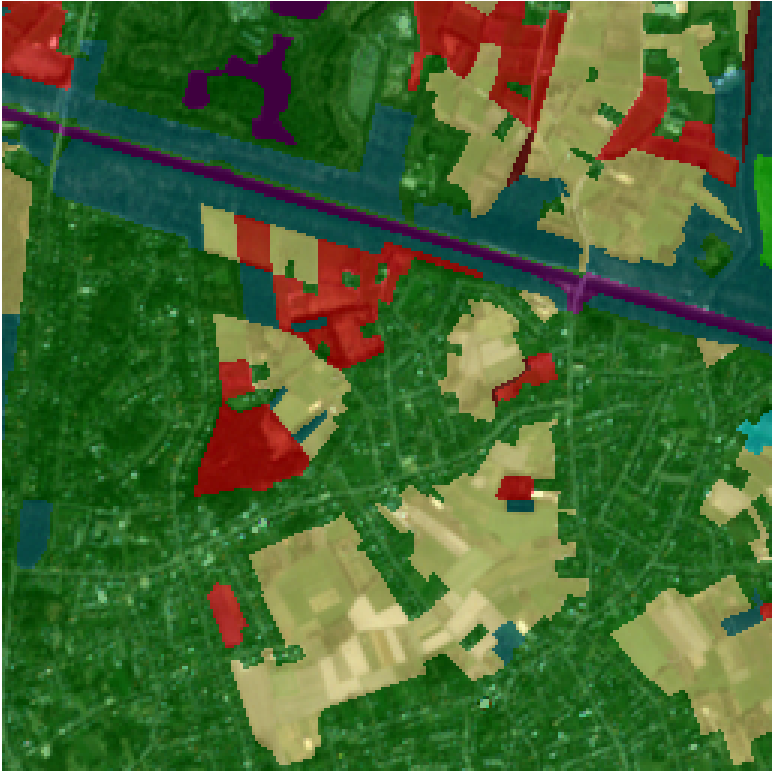} & \includegraphics[width=1.6in]{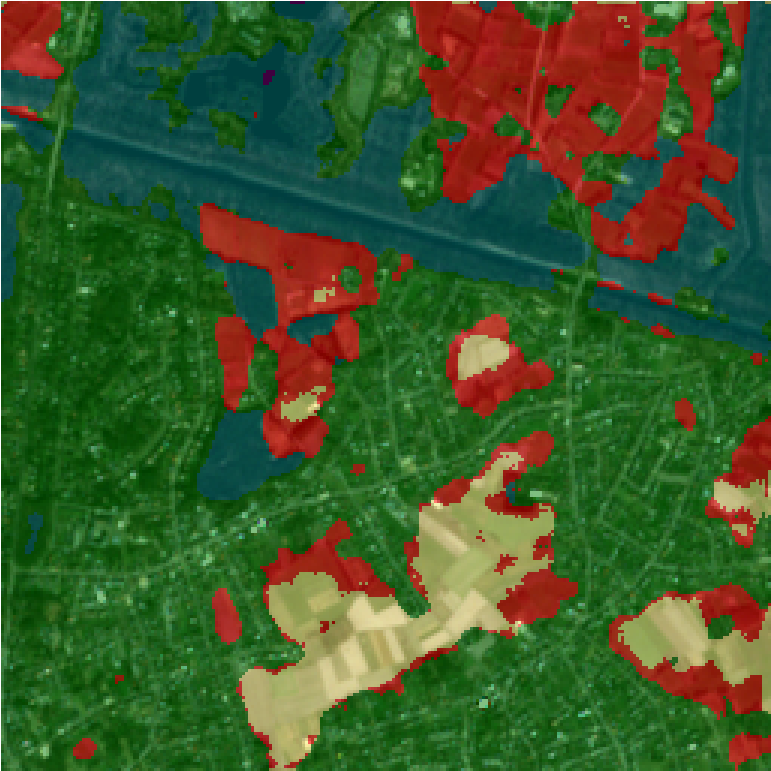} \\
    (a) Ground Truth & (b) ResNet2800M \\
    \includegraphics[width=1.6in]{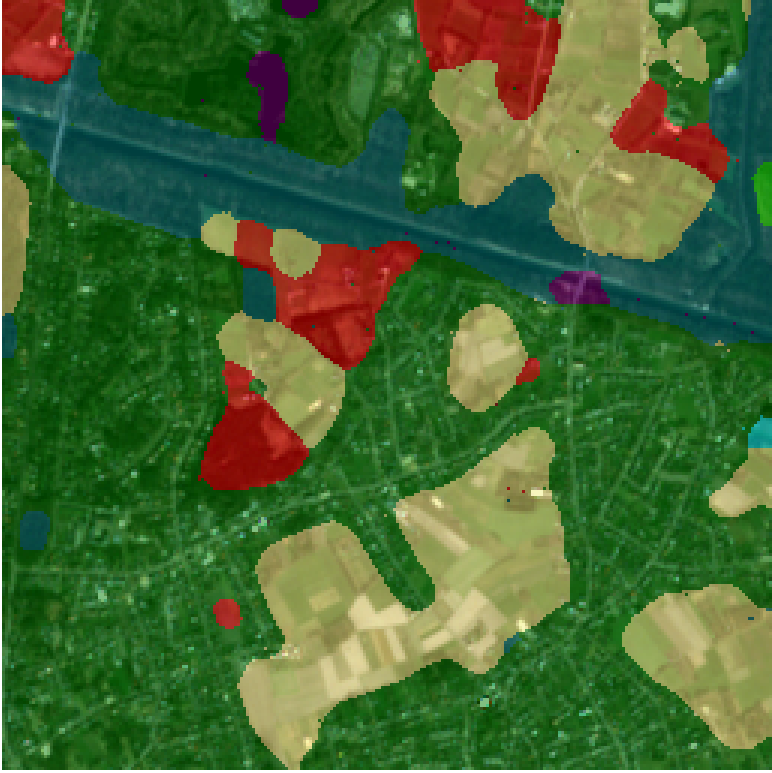} & \includegraphics[width=1.6in]{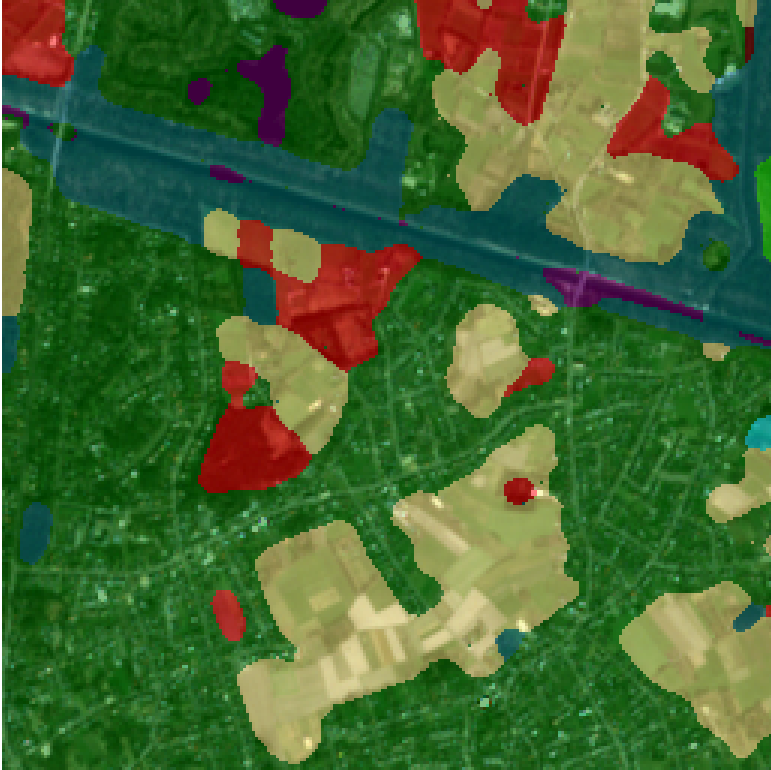} \\
    (c) Swin-h & (d) ChromaFormer-h \\
\end{tabular}
\caption{Demonstrative images: (a) Ground truth, (b) ResNet2800M, (c) Swin-h, and (d) ChromaFormer-h.}
\label{fig:comparison}
\end{figure}

Between the Swin and ChromaFormer models, the ChromaFormer models offer additional advantages, making them a better choice for multi-spectral segmentation tasks. The integration of the Spectral Dependency Module (SDM) into the Swin Transformer architecture allows ChromaFormer models to capture spectral dependencies more effectively, leading to higher accuracies without a significant increase in parameters or loss of scaling efficiency. For example, ChromaFormer achieves an accuracy of 92.25\%, approximately 0.91\% higher than Swint, with almost identical scaling coefficients. This demonstrates that ChromaFormer models enhance spectral feature learning while maintaining efficient scalability. Therefore, the ChromaFormer models are optimal for multi-spectral tasks requiring high performance and scalability, as they provide superior accuracy and maintain scaling efficiency compared to both their Swin counterparts and conventional models like ResNet and UNet++. \autoref{fig:comparison} shows some results of output from different models, as we can see from the demonstration ChromaFormer is doing better than ResNet and Swin Transformers in predicting minor classes.

\begin{figure}[htbp]
\begin{center}
    \includegraphics[width=3in]{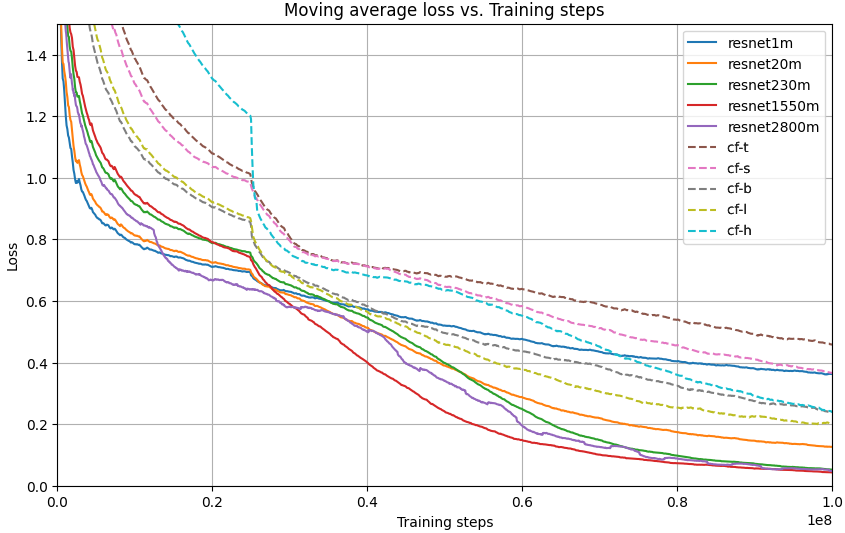}
    \includegraphics[width=3in]{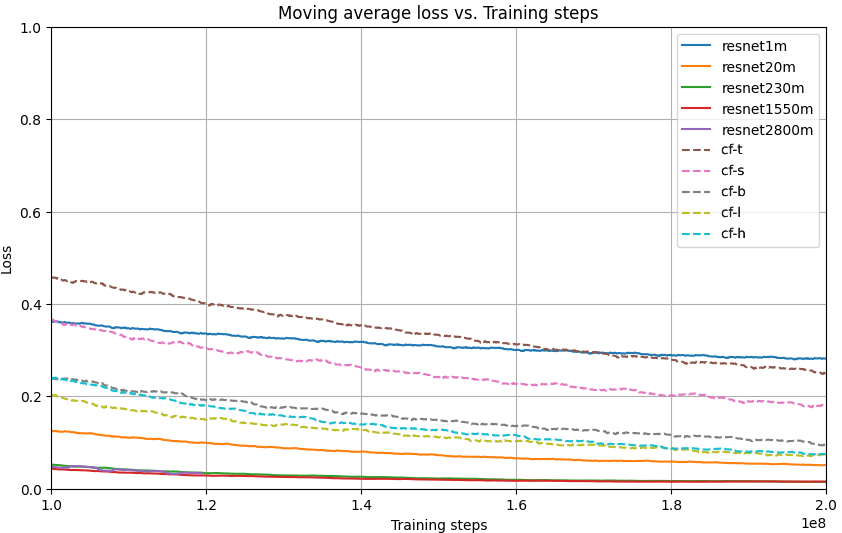}
\end{center}
\caption{Moving average loss curves of scaled networks, up: initial region of training, bottom: stable descending region}
\label{fig:loss_curves}
\end{figure}

\autoref{fig:loss_curves} shows the loss curve comparison between ChromaFormer and Resnet. The analysis of loss curves for various neural network models reveals distinct behaviors based on model size and architecture. Smaller ResNet models like ResNet1m and ResNet20m show rapid early loss descent but quickly reach saturation, indicating a capacity limitation in handling complex patterns as training progresses. In contrast, larger ResNet models and all ChromaFormer models demonstrate a more gradual loss reduction, suggesting that they can continue improving with extended training due to their greater capacity to learn complex data representations.

ChromaFormer models exhibit a slower but more persistent decline in loss, suggesting that they benefit from extended training epochs before showing signs of saturation, unlike conventional models like ResNet. This prolonged effectiveness in learning indicates that ChromaFormers, with their MSA and SDM modules, are capable of exploiting their architectural efficiency to handle complex data relationships over longer periods. While larger ResNet models tend to plateau earlier, larger ChromaFormer models, like ChromaFormer-l and ChromaFormer-h, continue to show potential for improvement well beyond the typical saturation points of conventional architectures, highlighting the distinct advantage of transformer-based models in sustained learning capability. We can also observe the scaling stops being rewarding for ResNet at around 1e8 sample passes for the loss gap between ResNet1550m and ResNet2800m is not distinct anymore while the loss ChromaFormer-l and ChromaFormer-h only start to converge at around 2e8 sample passes. This behavior underscores the fact that transformer models, despite their size, are inherently designed to scale more effectively with increasing data and training duration before experiencing diminishing returns.

\begin{figure}[htbp]
\begin{center}
    \includegraphics[width=3in]{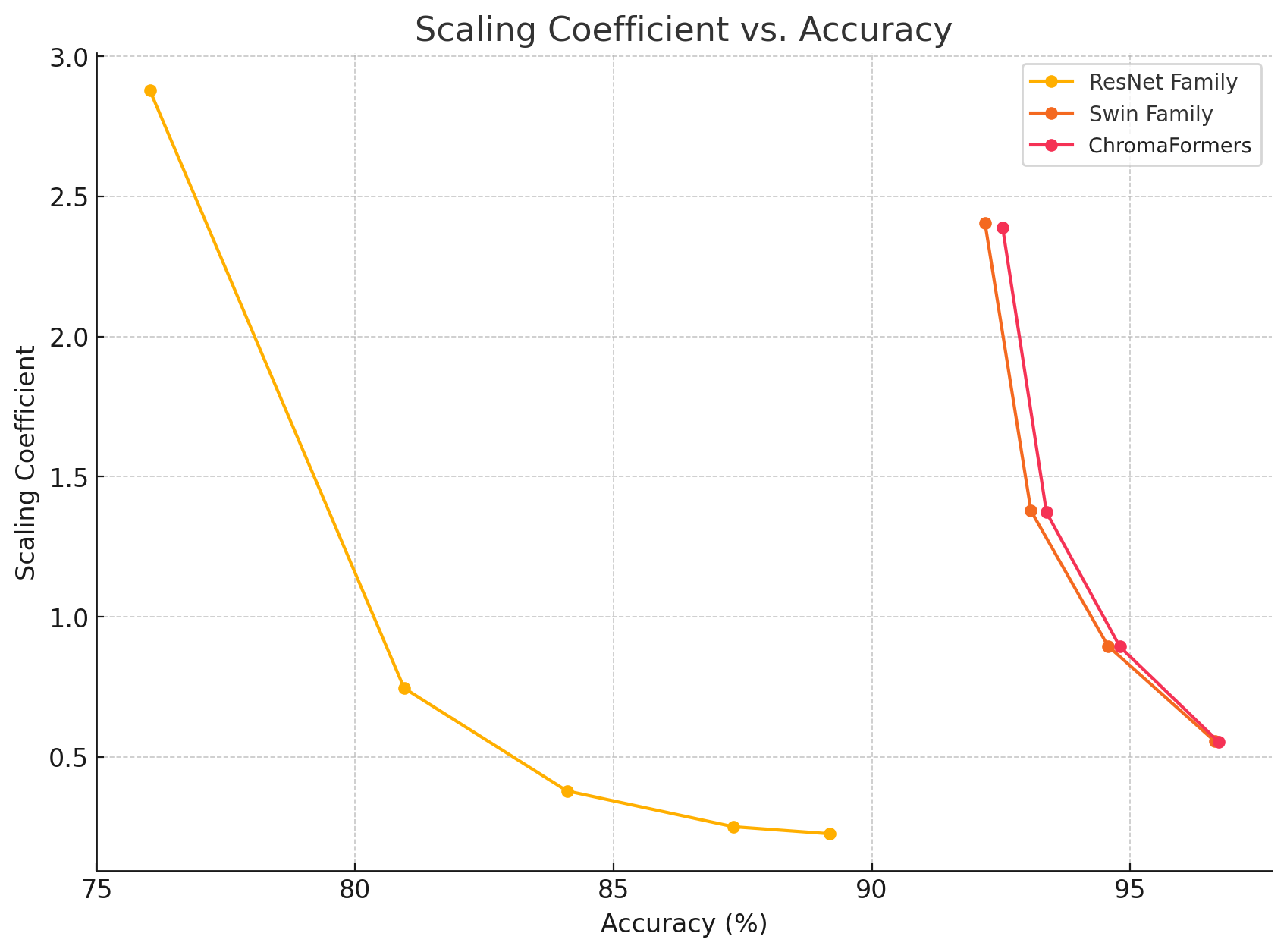}
    \includegraphics[width=3in]{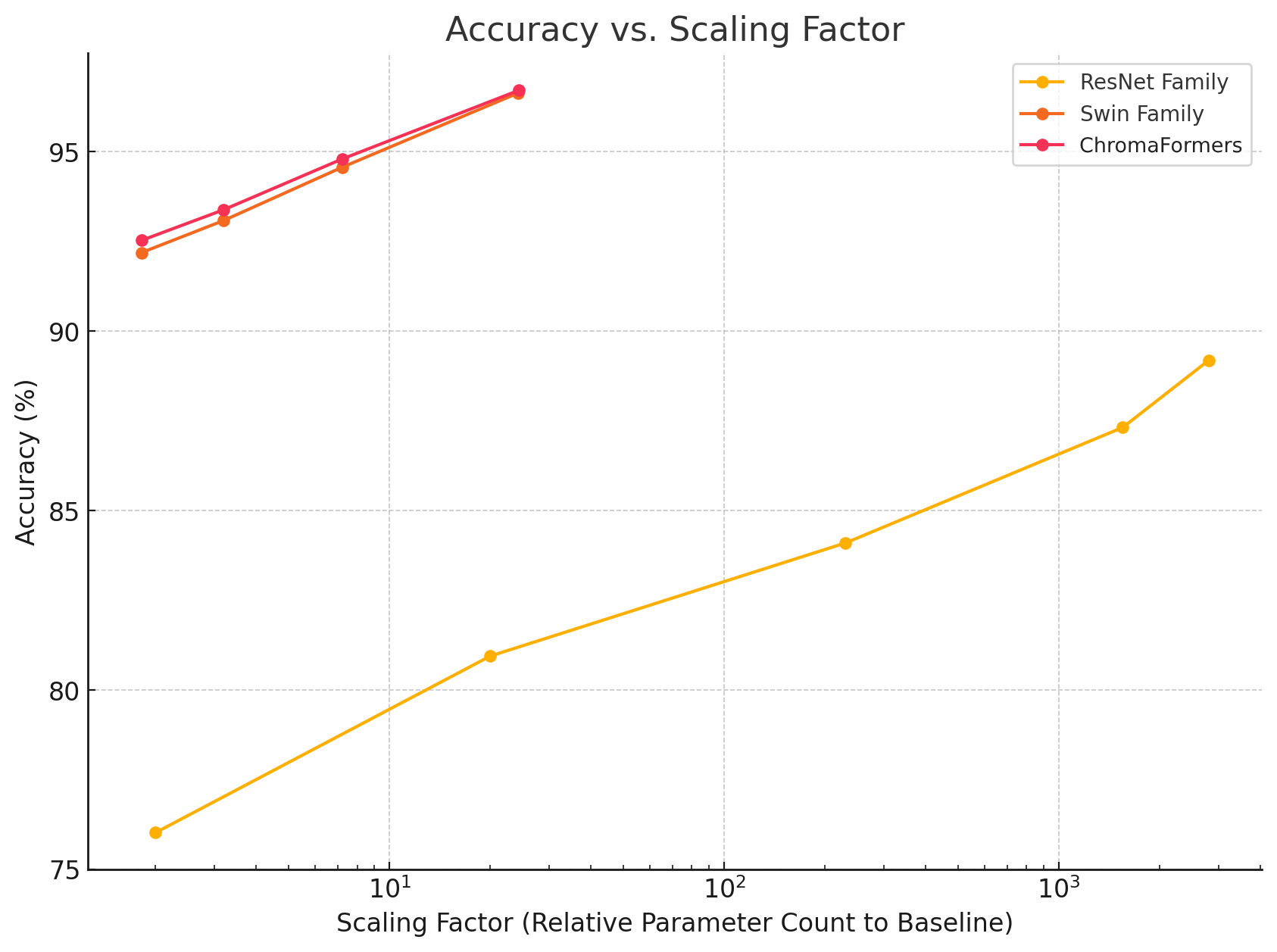}
\end{center}
\caption{Scaling properties of two architectural families. The "scaling coefficient" indicates how efficiently the network architecture scales. "Scaling factor" represents how many times the current model is larger in size compared to the baseline model. A higher scaling coefficient implies better scalability of the model, meaning the model does not suffer significantly from diminishing returns at that scaling step.}
\label{fig:scale_prop}
\end{figure}

\autoref{fig:scale_prop} shows comparative analysis of ChromaFormer, Swin, and ResNet model families. ChromaFormer models not only consistently maintain higher accuracy across all scaling scales but also demonstrate superior scalability. Both ChromaFormer and Swin showed much higher scaling efficiency than ResNet at the same accuracy level, and the ChromaFormer model has even higher scaling efficiency than the Swin model. This indicates that ChromaFormer only suffers from diminishing return at very high accuracy levels and at very large scales. Although data on larger ChromaFormer models is limited due to hardware constraints, it is projected that their scaling properties would outperform those of ResNet.

\begin{table*}[htbp]
\centering
\caption{Overall accuracy (\%) of models trained on different percentages of the training dataset. All models were validated on the full test set.}
\begin{tabular}{|l|c|c|c|c|c|}
\hline
\textbf{Model} & \textbf{5\% Data} & \textbf{10\% Data} & \textbf{30\% Data} & \textbf{50\% Data} & \textbf{100\% Data} \\ \hline
\multicolumn{6}{|c|}{\textbf{Small Models (\textasciitilde1M to 30M Parameters)}} \\ \hline
ResNet-1M      & 59.40 $\pm$ 0.10 & 61.07 $\pm$ 0.07 & 71.00 $\pm$ 0.04 & 74.52 $\pm$ 0.03 & 75.92 $\pm$ 0.02 \\ \hline
ResNet-2M      & 58.33 $\pm$ 0.10 & 63.14 $\pm$ 0.07 & 71.30 $\pm$ 0.04 & 74.35 $\pm$ 0.03 & 76.03 $\pm$ 0.02 \\ \hline
UNet++         & 60.47 $\pm$ 0.10 & 61.11 $\pm$ 0.07 & 62.38 $\pm$ 0.04 & 63.61 $\pm$ 0.03 & 64.48 $\pm$ 0.02 \\ \hline
ResNet-20M     & 58.61 $\pm$ 0.10 & 63.35 $\pm$ 0.07 & 75.20 $\pm$ 0.04 & 79.08 $\pm$ 0.03 & 80.95 $\pm$ 0.02 \\ \hline
Swint          & 62.05 $\pm$ 0.10 & 67.35 $\pm$ 0.07 & 79.10 $\pm$ 0.03 & 85.40 $\pm$ 0.02 & 91.34 $\pm$ 0.01 \\ \hline
ChromaFormer-t & 64.02 $\pm$ 0.10 & 69.51 $\pm$ 0.07 & 81.33 $\pm$ 0.03 & 87.33 $\pm$ 0.02 & 92.25 $\pm$ 0.01 \\ \hline
\multicolumn{6}{|c|}{\textbf{Medium Models (\textasciitilde50M to 100M Parameters)}} \\ \hline
ResNet-230M    & 61.08 $\pm$ 0.10 & 67.99 $\pm$ 0.07 & 78.16 $\pm$ 0.03 & 82.45 $\pm$ 0.02 & 84.10 $\pm$ 0.02 \\ \hline
Swins          & 62.65 $\pm$ 0.10 & 68.65 $\pm$ 0.07 & 80.25 $\pm$ 0.03 & 86.20 $\pm$ 0.02 & 92.19 $\pm$ 0.01 \\ \hline
ChromaFormer-s & 64.06 $\pm$ 0.10 & 70.02 $\pm$ 0.07 & 81.91 $\pm$ 0.03 & 87.66 $\pm$ 0.02 & 92.53 $\pm$ 0.01 \\ \hline
Swinb          & 62.80 $\pm$ 0.10 & 65.90 $\pm$ 0.07 & 80.60 $\pm$ 0.03 & 86.70 $\pm$ 0.02 & 93.08 $\pm$ 0.01 \\ \hline
ChromaFormer-b & 63.71 $\pm$ 0.10 & 66.80 $\pm$ 0.07 & 81.56 $\pm$ 0.03 & 87.67 $\pm$ 0.02 & 93.38 $\pm$ 0.01 \\ \hline
\multicolumn{6}{|c|}{\textbf{Large Models (\textasciitilde150M to 300M Parameters)}} \\ \hline
ResNet-1550M   & 59.95 $\pm$ 0.10 & 64.65 $\pm$ 0.07 & 79.99 $\pm$ 0.04 & 85.20 $\pm$ 0.03 & 87.32 $\pm$ 0.02 \\ \hline
Swinl          & 65.70 $\pm$ 0.09 & 69.90 $\pm$ 0.06 & 83.65 $\pm$ 0.03 & 89.35 $\pm$ 0.02 & 94.57 $\pm$ 0.01 \\ \hline
ChromaFormer-l & 66.09 $\pm$ 0.10 & 70.36 $\pm$ 0.07 & 84.13 $\pm$ 0.03 & 89.80 $\pm$ 0.02 & 94.80 $\pm$ 0.01 \\ \hline
\multicolumn{6}{|c|}{\textbf{Extra-Large Models (\textasciitilde650M to 2800M Parameters)}} \\ \hline
ResNet-2800M   & 60.19 $\pm$ 0.10 & 67.59 $\pm$ 0.07 & 80.08 $\pm$ 0.04 & 85.34 $\pm$ 0.02 & 89.19 $\pm$ 0.01 \\ \hline
Swinh          & 63.12 $\pm$ 0.10 & 70.53 $\pm$ 0.07 & 84.27 $\pm$ 0.03 & 91.08 $\pm$ 0.02 & 96.64 $\pm$ 0.01 \\ \hline
ChromaFormer-h & 63.66 $\pm$ 0.10 & 70.60 $\pm$ 0.07 & 84.45 $\pm$ 0.03 & 91.19 $\pm$ 0.02 & 96.71 $\pm$ 0.01 \\ \hline
\end{tabular}
\label{tab:accuracy-noisy}
\end{table*}

\begin{figure*}[htbp]
\begin{center}
    \includegraphics[width=6in]{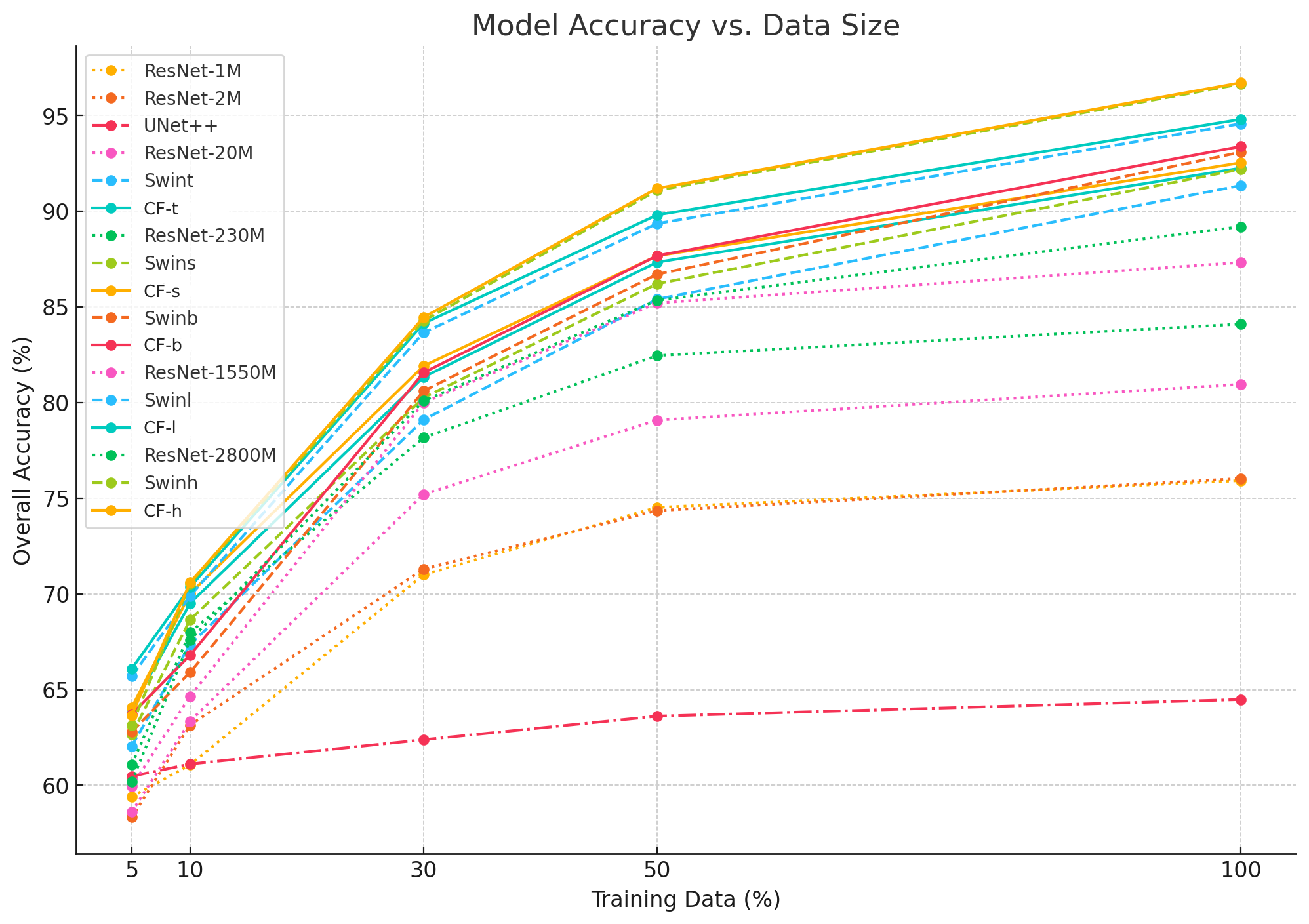}
\end{center}
\caption{The accuracy curves of models vs. training data size}
\label{fig:acc_wt_datasize}
\end{figure*}

As \autoref{tab:accuracy-noisy} and \autoref{fig:acc_wt_datasize} show, when exposed to a small amount of data, small models, which have fewer parameters, are more appropriate for scenarios with limited data. The small models are significantly more energy efficient than their larger counterparts, with virtually the same level of accuracy scores. These models tend to generalize better and are less likely to overfit when data is scarce. As the dataset size grows, larger models begin to show their advantages. With more data, these models can learn more complex patterns, which is evident from the performance jump observed in large and extra-large models with increased training data.

The impact of increasing model size becomes more obvious with larger datasets. Smaller models reach their capacity earlier, while larger models continue to improve. We can also infer from the \autoref{fig:acc_wt_datasize} that there is still room for further increase in ChromaFormer and Swin models, demonstrating that at this dataset size scale, it is significantly more important to scale up model size than to optimize the architectural design of the network since even conventional architectural choices such as ResNet can perform much better than using smaller variations.

\begin{table*}[ht]
\centering
\caption{Confusion matrix for 14 classes with ChromaFormer-h on test dataset with 5 million input samples. 
CDH = Coastal dune habitats, 
CL = Cultivated land, 
G = Grasslands, 
H = Heathland, 
IM = Inland marshes, 
MH = Marine habitats, 
PV = Pioneer vegetation, 
SLF-ns = Small landscape features (not specified), 
SNWLF = Small non-woody landscape features, 
SWLF = Small woody landscape features, 
U = Unknown, 
UA = Urban areas, 
WB = Water bodies, 
WS = Woodland and shrub.}
\resizebox{\textwidth}{!}{%
\begin{tabular}{l|rrrrrrrrrrrrrr}
\hline
 & \textbf{CDH} & \textbf{CL} & \textbf{G} & \textbf{H} & \textbf{IM} & \textbf{MH} & \textbf{PV} & \textbf{SLF-ns} & \textbf{SNWLF} & \textbf{SWLF} & \textbf{U} & \textbf{UA} & \textbf{WB} & \textbf{WS} \\
\hline
\textbf{CDH}    & 18599 & 340    & 750    & 11    & 7     & 450   & 90    & 0      & 0      & 5      & 0     & 484   & 101   & 137   \\
\textbf{CL}     & 15    & 1775401 & 59532  & 138   & 78    & 158   & 192   & 7      & 30     & 347    & 1     & 4608  & 704   & 2085  \\
\textbf{G}      & 3     & 48859  & 1245308 & 53    & 89    & 21    & 175   & 6      & 18     & 348    & 0     & 5433  & 292   & 2237  \\
\textbf{H}      & 1     & 102    & 314    & 21443 & 28    & 0     & 26    & 0      & 0      & 5      & 5     & 69    & 26    & 620   \\
\textbf{IM}     & 0     & 59     & 392    & 0     & 19147 & 10    & 14    & 0      & 1      & 6      & 0     & 53    & 58    & 171   \\
\textbf{MH}     & 155   & 134    & 140    & 2     & 17    & 45001 & 24    & 0      & 0      & 1      & 0     & 495   & 12302 & 40    \\
\textbf{PV}     & 1     & 267    & 546    & 5     & 9     & 3     & 42299 & 0      & 2      & 12     & 0     & 696   & 54    & 206   \\
\textbf{SLF-ns} & 0     & 44     & 16     & 0     & 0     & 0     & 0     & 1383   & 1      & 0      & 0     & 4     & 0     & 2     \\
\textbf{SNWLF}  & 0     & 371    & 301    & 0     & 2     & 0     & 2     & 0      & 16007  & 12     & 0     & 109   & 20    & 60    \\
\textbf{SWLF}   & 0     & 406    & 655    & 2     & 5     & 0     & 9     & 0      & 1      & 37684  & 0     & 235   & 27    & 371   \\
\textbf{U}      & 0     & 0      & 0      & 0     & 0     & 0     & 0     & 0      & 0      & 0      & 25    & 0     & 0     & 4     \\
\textbf{UA}     & 22    & 2283   & 4441   & 27    & 24    & 27    & 160   & 6      & 13     & 125    & 0     & 1105916 & 407  & 1920  \\
\textbf{WB}     & 0     & 183    & 498    & 18    & 26    & 132   & 18    & 0      & 1      & 18     & 0     & 353   & 85034 & 140   \\
\textbf{WS}     & 2     & 936    & 4204   & 174   & 108   & 0     & 124   & 5      & 5      & 188    & 2     & 1704  & 165   & 430760\\
\hline
\end{tabular}}
\label{tab:conf_matrix}
\end{table*}

\section{Conclusions and limitations}

In this paper, we have explored the scaling laws governing neural networks applied to large-scale, multi-spectral remote sensing tasks. Our extensive experiments on the Biological Valuation Map (BVM) of Flanders—a densely labeled, large-scale dataset—demonstrated that larger models with appropriate architectural choices significantly outperform traditional architectures like UNet++ and ResNet. By introducing a Spectral Dependency Module (SDM) integrated into a Swin Transformer architecture, we developed ChromaFormer, a multi-spectral Transformer, that effectively captures spectral dependencies in multi-spectral data. Specifically, we showed that our models achieve superior accuracy and scaling efficiency, with models containing hundreds of millions of parameters yielding over 96\% accuracy, while smaller conventional models lag behind. These findings underscore the importance of matching model complexity to dataset scale and leveraging advanced architectural designs to harness the full potential of large multi-spectral datasets.

Despite the promising results, our study has several limitations. Firstly, our experiments were conducted solely on the BVM dataset, which is restricted to the Belgian Flemish region. To generalize our findings, it is is of interest to test the proposed models on a broader range of land regions in different countries, leveraging multi-spectral datasets of similar scale, which governments and organizations are increasingly capable of providing. Secondly, due to computational resource constraints, we were unable to further scale up the size of the ChromaFormer models. Future work could explore even larger models, potentially requiring inter-node training setups, although this would introduce additional computational overhead that must be carefully managed. Lastly, while we focused on comparing our models with standard architectures like ResNet and UNet++, there exists a wide variety of state-of-the-art models, including other large transformers and conventional models. Incorporating and testing the Spectral Dependency Module within these architectures could provide deeper insights into their scaling behaviors and further consolidate the configurations of SDM. The code is available at: \url{https://github.com/limingshi1994/ChromaFormer}

\section*{Acknowledgments}

We acknowledge funding from the Flemish Government
under FWO projects GEO.INFORMED S001421N and G0G2921N, and the Onderzoeksprogramma Artifici\"{e}le Intelligentie
(AI) Vlaanderen programme

\bibliography{ieeetgrs}

\begin{thebibliography}{10}
\providecommand{\url}[1]{#1}
\csname url@samestyle\endcsname
\providecommand{\newblock}{\relax}
\providecommand{\bibinfo}[2]{#2}
\providecommand{\BIBentrySTDinterwordspacing}{\spaceskip=0pt\relax}
\providecommand{\BIBentryALTinterwordstretchfactor}{4}
\providecommand{\BIBentryALTinterwordspacing}{\spaceskip=\fontdimen2\font plus
\BIBentryALTinterwordstretchfactor\fontdimen3\font minus \fontdimen4\font\relax}
\providecommand{\BIBforeignlanguage}[2]{{%
\expandafter\ifx\csname l@#1\endcsname\relax
\typeout{** WARNING: IEEEtran.bst: No hyphenation pattern has been}%
\typeout{** loaded for the language `#1'. Using the pattern for}%
\typeout{** the default language instead.}%
\else
\language=\csname l@#1\endcsname
\fi
#2}}
\providecommand{\BIBdecl}{\relax}
\BIBdecl

\bibitem{zhang2016deep}
L.~Zhang, L.~Zhang, and B.~Du, ``Deep learning for remote sensing data: A technical tutorial on the state of the art,'' \emph{IEEE Geoscience and Remote Sensing Magazine}, vol.~4, no.~2, pp. 22--40, 2016.

\bibitem{li2019deep}
Y.~Li, H.~Zhang, X.~Xue, Y.~Jiang, and Q.~Shen, ``Deep learning for remote sensing image classification: A survey,'' \emph{Wiley Interdisciplinary Reviews: Data Mining and Knowledge Discovery}, vol.~8, no.~6, p. e1264, 2018.

\bibitem{huhyper}
\BIBentryALTinterwordspacing
W.~Hu, Y.~Huang, L.~Wei, F.~Zhang, and H.~Li, ``Deep convolutional neural networks for hyperspectral image classification,'' \emph{Journal of Sensors}, vol. 2015, no.~1, p. 258619, 2015. [Online]. Available: \url{https://onlinelibrary.wiley.com/doi/abs/10.1155/2015/258619}
\BIBentrySTDinterwordspacing

\bibitem{vaswani2017attention}
\BIBentryALTinterwordspacing
A.~Vaswani, N.~Shazeer, N.~Parmar, J.~Uszkoreit, L.~Jones, A.~N. Gomez, L.~u. Kaiser, and I.~Polosukhin, ``Attention is all you need,'' in \emph{Advances in Neural Information Processing Systems}, I.~Guyon, U.~V. Luxburg, S.~Bengio, H.~Wallach, R.~Fergus, S.~Vishwanathan, and R.~Garnett, Eds., vol.~30.\hskip 1em plus 0.5em minus 0.4em\relax Curran Associates, Inc., 2017. [Online]. Available: \url{https://proceedings.neurips.cc/paper_files/paper/2017/file/3f5ee243547dee91fbd053c1c4a845aa-Paper.pdf}
\BIBentrySTDinterwordspacing

\bibitem{carion2020end}
N.~Carion, F.~Massa, G.~Synnaeve, N.~Usunier, A.~Kirillov, and S.~Zagoruyko, ``End-to-end object detection with transformers,'' in \emph{European conference on computer vision}.\hskip 1em plus 0.5em minus 0.4em\relax Springer, 2020, pp. 213--229.

\bibitem{han2022survey}
K.~Han, Y.~Wang, H.~Chen, X.~Chen, J.~Guo, Z.~Liu, Y.~Tang, A.~Xiao, C.~Xu, Y.~Xu \emph{et~al.}, ``A survey on vision transformer,'' \emph{IEEE transactions on pattern analysis and machine intelligence}, vol.~45, no.~1, pp. 87--110, 2022.

\bibitem{dosovitskiy2020image}
\BIBentryALTinterwordspacing
A.~Dosovitskiy, L.~Beyer, A.~Kolesnikov, D.~Weissenborn, X.~Zhai, T.~Unterthiner, M.~Dehghani, M.~Minderer, G.~Heigold, S.~Gelly, J.~Uszkoreit, and N.~Houlsby, ``An image is worth 16x16 words: Transformers for image recognition at scale,'' in \emph{International Conference on Learning Representations}, 2021. [Online]. Available: \url{https://openreview.net/forum?id=YicbFdNTTy}
\BIBentrySTDinterwordspacing

\bibitem{liu2021swin}
Z.~Liu, Y.~Lin, Y.~Cao, H.~Hu, Y.~Wei, Z.~Zhang, S.~Lin, and B.~Guo, ``Swin transformer: Hierarchical vision transformer using shifted windows,'' in \emph{Proceedings of the IEEE/CVF international conference on computer vision}, 2021, pp. 10\,012--10\,022.

\bibitem{stunet}
X.~He, Y.~Zhou, J.~Zhao, D.~Zhang, R.~Yao, and Y.~Xue, ``Swin transformer embedding unet for remote sensing image semantic segmentation,'' \emph{IEEE Transactions on Geoscience and Remote Sensing}, vol.~60, pp. 1--15, 2022.

\bibitem{DBLP:journals/corr/abs-2001-08361}
\BIBentryALTinterwordspacing
J.~Kaplan, S.~McCandlish, T.~Henighan, T.~B. Brown, B.~Chess, R.~Child, S.~Gray, A.~Radford, J.~Wu, and D.~Amodei, ``Scaling laws for neural language models,'' \emph{CoRR}, vol. abs/2001.08361, 2020. [Online]. Available: \url{https://arxiv.org/abs/2001.08361}
\BIBentrySTDinterwordspacing

\bibitem{zhureview}
X.~X. Zhu, D.~Tuia, L.~Mou, G.-S. Xia, L.~Zhang, F.~Xu, and F.~Fraundorfer, ``Deep learning in remote sensing: A comprehensive review and list of resources,'' \emph{IEEE Geoscience and Remote Sensing Magazine}, vol.~5, no.~4, pp. 8--36, 2017.

\bibitem{maggiori2016convolutional}
E.~Maggiori, Y.~Tarabalka, G.~Charpiat, and P.~Alliez, ``Convolutional neural networks for large-scale remote-sensing image classification,'' \emph{IEEE Transactions on geoscience and remote sensing}, vol.~55, no.~2, pp. 645--657, 2016.

\bibitem{transsurvey}
\BIBentryALTinterwordspacing
A.~A. Aleissaee, A.~Kumar, R.~M. Anwer, S.~Khan, H.~Cholakkal, G.-S. Xia, and F.~S. Khan, ``Transformers in remote sensing: A survey,'' \emph{Remote Sensing}, vol.~15, no.~7, 2023. [Online]. Available: \url{https://www.mdpi.com/2072-4292/15/7/1860}
\BIBentrySTDinterwordspacing

\bibitem{hangssatt}
R.~Hang, Z.~Li, Q.~Liu, P.~Ghamisi, and S.~S. Bhattacharyya, ``Hyperspectral image classification with attention-aided cnns,'' \emph{IEEE Transactions on Geoscience and Remote Sensing}, vol.~59, no.~3, pp. 2281--2293, 2021.

\bibitem{resssatt}
S.~K. Roy, S.~Manna, T.~Song, and L.~Bruzzone, ``Attention-based adaptive spectral–spatial kernel resnet for hyperspectral image classification,'' \emph{IEEE Transactions on Geoscience and Remote Sensing}, vol.~59, no.~9, pp. 7831--7843, 2021.

\bibitem{fas}
Z.~Zhong, Y.~Li, L.~Ma, J.~Li, and W.-S. Zheng, ``Spectral–spatial transformer network for hyperspectral image classification: A factorized architecture search framework,'' \emph{IEEE Transactions on Geoscience and Remote Sensing}, vol.~60, pp. 1--15, 2022.

\bibitem{bvm}
M.~Li, D.~Grujicic, S.~De~Saeger, S.~Heremans, B.~Somers, and M.~B. Blaschko, ``Biological valuation map of {Flanders}: A {Sentinel-2} imagery analysis,'' in \emph{IEEE International Geoscience and Remote Sensing Symposium}, 2024.

\bibitem{bergamasco2023dual}
L.~Bergamasco, F.~Bovolo, and L.~Bruzzone, ``A dual-branch deep learning architecture for multisensor and multitemporal remote sensing semantic segmentation,'' \emph{IEEE Journal of Selected Topics in Applied Earth Observations and Remote Sensing}, vol.~16, pp. 2147--2162, 2023.

\bibitem{peng2023rsbnet}
C.~Peng, Y.~Li, R.~Shang, and L.~Jiao, ``Rsbnet: One-shot neural architecture search for a backbone network in remote sensing image recognition,'' \emph{Neurocomputing}, vol. 537, pp. 110--127, 2023.

\bibitem{roy2023multimodal}
S.~K. Roy, A.~Deria, D.~Hong, B.~Rasti, A.~Plaza, and J.~Chanussot, ``Multimodal fusion transformer for remote sensing image classification,'' \emph{IEEE Transactions on Geoscience and Remote Sensing}, vol.~61, pp. 1--20, 2023.

\bibitem{yuan2023litest}
W.~Yuan, X.~Zhang, J.~Shi, and J.~Wang, ``Litest-net: a hybrid model of lite swin transformer and convolution for building extraction from remote sensing image,'' \emph{Remote Sensing}, vol.~15, no.~8, p. 1996, 2023.

\bibitem{lv2023shapeformer}
P.~Lv, L.~Ma, Q.~Li, and F.~Du, ``Shapeformer: A shape-enhanced vision transformer model for optical remote sensing image landslide detection,'' \emph{IEEE Journal of Selected Topics in Applied Earth Observations and Remote Sensing}, vol.~16, pp. 2681--2689, 2023.

\bibitem{zhang2023efficient}
C.~Zhang, J.~Su, Y.~Ju, K.-M. Lam, and Q.~Wang, ``Efficient inductive vision transformer for oriented object detection in remote sensing imagery,'' \emph{IEEE Transactions on Geoscience and Remote Sensing}, 2023.

\bibitem{salinas}
\BIBentryALTinterwordspacing
M.~Graña, M.~Veganzons, and B.~Ayerdi, ``Hyperspectral remote sensing scenes,'' 2021. [Online]. Available: \url{https://www.ehu.eus/ccwintco/index.php/Hyperspectral_Remote_Sensing_Scenes}
\BIBentrySTDinterwordspacing

\bibitem{clasen2024reben}
K.~N. Clasen, L.~Hackel, T.~Burgert, G.~Sumbul, B.~Demir, and V.~Markl, ``{reBEN}: Refined {BigEarthNet} dataset for remote sensing image analysis,'' \emph{arXiv preprint arXiv:2407.03653}, 2024.

\bibitem{li2022}
\BIBentryALTinterwordspacing
J.~Li, H.~Wang, A.~Zhang, and Y.~Liu, ``Semantic segmentation of hyperspectral remote sensing images based on pse-unet model,'' \emph{Sensors}, vol.~22, no.~24, 2022. [Online]. Available: \url{https://www.mdpi.com/1424-8220/22/24/9678}
\BIBentrySTDinterwordspacing

\bibitem{3dcnn}
\BIBentryALTinterwordspacing
L.~Liu, E.~M. Awwad, Y.~A. Ali, M.~Al-Razgan, A.~Maarouf, L.~Abualigah, and A.~N. Hoshyar, ``Multi-dataset hyper-cnn for hyperspectral image segmentation of remote sensing images,'' \emph{Processes}, vol.~11, no.~2, 2023. [Online]. Available: \url{https://www.mdpi.com/2227-9717/11/2/435}
\BIBentrySTDinterwordspacing

\bibitem{hybridsn}
S.~K. Roy, G.~Krishna, S.~R. Dubey, and B.~B. Chaudhuri, ``Hybridsn: Exploring 3-d–2-d cnn feature hierarchy for hyperspectral image classification,'' \emph{IEEE Geoscience and Remote Sensing Letters}, vol.~17, no.~2, pp. 277--281, 2020.

\bibitem{smale}
\BIBentryALTinterwordspacing
N.~Liao, J.~Gong, W.~Li, C.~Li, C.~Zhang, and B.~Guo, ``Smale: Hyperspectral image classification via superpixels and manifold learning,'' \emph{Remote Sensing}, vol.~16, no.~18, 2024. [Online]. Available: \url{https://www.mdpi.com/2072-4292/16/18/3442}
\BIBentrySTDinterwordspacing

\bibitem{agos}
Q.~Bi, B.~Zhou, K.~Qin, Q.~Ye, and G.-S. Xia, ``All grains, one scheme (agos): Learning multigrain instance representation for aerial scene classification,'' \emph{IEEE Transactions on Geoscience and Remote Sensing}, vol.~60, pp. 1--17, 2022.

\bibitem{ms2ap}
Q.~Bi, H.~Zhang, and K.~Qin, ``Multi-scale stacking attention pooling for remote sensing scene classification,'' \emph{Neurocomputing}, vol. 436, pp. 147--161, 2021.

\bibitem{lsenet}
Q.~Bi, K.~Qin, H.~Zhang, and G.-S. Xia, ``Local semantic enhanced convnet for aerial scene recognition,'' \emph{IEEE Transactions on Image Processing}, vol.~30, pp. 6498--6511, 2021.

\bibitem{vggvd16}
F.~{\"O}zyurt, E.~Ava, and E.~Sert, ``Uc-merced image classification with cnn feature reduction using wavelet entropy optimized with genetic algorithm,'' \emph{Traitement du Signal}, 2020.

\bibitem{pgnet}
\BIBentryALTinterwordspacing
B.~Liu, J.~Hu, X.~Bi, W.~Li, and X.~Gao, ``Pgnet: Positioning guidance network for semantic segmentation of very-high-resolution remote sensing images,'' \emph{Remote Sensing}, vol.~14, no.~17, 2022. [Online]. Available: \url{https://www.mdpi.com/2072-4292/14/17/4219}
\BIBentrySTDinterwordspacing

\bibitem{manet}
R.~Li, S.~Zheng, C.~Zhang, C.~Duan, J.~Su, L.~Wang, and P.~M. Atkinson, ``Multiattention network for semantic segmentation of fine-resolution remote sensing images,'' \emph{IEEE Transactions on Geoscience and Remote Sensing}, vol.~60, pp. 1--13, 2021.

\bibitem{emnet}
X.~Li, Y.~Li, J.~Ai, Z.~Shu, J.~Xia, and Y.~Xia, ``Semantic segmentation of uav remote sensing images based on edge feature fusing and multi-level upsampling integrated with deeplabv3+,'' \emph{Plos one}, vol.~18, no.~1, p. e0279097, 2023.

\bibitem{deeplabv3}
L.-C. Chen, Y.~Zhu, G.~Papandreou, F.~Schroff, and H.~Adam, ``Encoder-decoder with atrous separable convolution for semantic image segmentation,'' in \emph{Proceedings of the European conference on computer vision (ECCV)}, 2018, pp. 801--818.

\bibitem{cmunet}
\BIBentryALTinterwordspacing
M.~Liu, J.~Dan, Z.~Lu, Y.~Yu, Y.~Li, and X.~Li, ``Cm-unet: Hybrid cnn-mamba unet for remote sensing image semantic segmentation,'' 2024. [Online]. Available: \url{https://arxiv.org/abs/2405.10530}
\BIBentrySTDinterwordspacing

\bibitem{sscnet}
\BIBentryALTinterwordspacing
X.~Li, F.~Xu, X.~Yong, D.~Chen, R.~Xia, B.~Ye, H.~Gao, Z.~Chen, and X.~Lyu, ``Sscnet: A spectrum-space collaborative network for semantic segmentation of remote sensing images,'' \emph{Remote Sensing}, vol.~15, no.~23, 2023. [Online]. Available: \url{https://www.mdpi.com/2072-4292/15/23/5610}
\BIBentrySTDinterwordspacing

\bibitem{hcanet}
X.~Li, F.~Xu, R.~Xia, X.~Lyu, H.~Gao, and Y.~Tong, ``Hybridizing cross-level contextual and attentive representations for remote sensing imagery semantic segmentation,'' \emph{Remote Sensing}, vol.~13, no.~15, p. 2986, 2021.

\bibitem{rs16162930}
\BIBentryALTinterwordspacing
T.~Hanyu, K.~Yamazaki, M.~Tran, R.~A. McCann, H.~Liao, C.~Rainwater, M.~Adkins, J.~Cothren, and N.~Le, ``Aerialformer: Multi-resolution transformer for aerial image segmentation,'' \emph{Remote Sensing}, vol.~16, no.~16, 2024. [Online]. Available: \url{https://www.mdpi.com/2072-4292/16/16/2930}
\BIBentrySTDinterwordspacing

\bibitem{dcswin}
L.~Wang, R.~Li, C.~Duan, C.~Zhang, X.~Meng, and S.~Fang, ``A novel transformer based semantic segmentation scheme for fine-resolution remote sensing images,'' \emph{IEEE Geoscience and Remote Sensing Letters}, vol.~19, pp. 1--5, 2022.

\bibitem{segnext}
M.-H. Guo, C.-Z. Lu, Q.~Hou, Z.~Liu, M.-M. Cheng, and S.-M. Hu, ``Segnext: Rethinking convolutional attention design for semantic segmentation,'' \emph{Advances in Neural Information Processing Systems}, vol.~35, pp. 1140--1156, 2022.

\bibitem{unetensemble}
I.~Dimitrovski, V.~Spasev, S.~Loshkovska, and I.~Kitanovski, ``U-net ensemble for enhanced semantic segmentation in remote sensing imagery,'' \emph{Remote Sensing}, vol.~16, no.~12, p. 2077, 2024.

\bibitem{sfanet}
G.~Hwang, J.~Jeong, and S.~J. Lee, ``Sfa-net: Semantic feature adjustment network for remote sensing image segmentation,'' \emph{Remote Sensing}, vol.~16, no.~17, p. 3278, 2024.

\bibitem{vitg12x4}
K.~Cha, J.~Seo, and T.~Lee, ``A billion-scale foundation model for remote sensing images,'' \emph{IEEE Journal of Selected Topics in Applied Earth Observations and Remote Sensing}, pp. 1--17, 2024.

\bibitem{lsknet}
\BIBentryALTinterwordspacing
Y.~Li, X.~Li, Y.~Dai, Q.~Hou, L.~Liu, Y.~Liu, M.-M. Cheng, and J.~Yang, ``Lsknet: A foundation lightweight backbone for remote sensing,'' \emph{International Journal of Computer Vision}, vol. 133, no.~3, pp. 1410--1431, Mar 2025. [Online]. Available: \url{https://doi.org/10.1007/s11263-024-02247-9}
\BIBentrySTDinterwordspacing

\bibitem{deeptrinet}
T.~B. Ovi, S.~Mosharrof, N.~Bashree, M.~N. Islam, and M.~S. Islam, ``Deeptrinet: A tri-level attention-based deeplabv3+ architecture for semantic segmentation of satellite images,'' in \emph{International Conference on Human-Centric Smart Computing}.\hskip 1em plus 0.5em minus 0.4em\relax Springer, 2023, pp. 373--384.

\bibitem{ssleo}
Y.~Wang, N.~A.~A. Braham, Z.~Xiong, C.~Liu, C.~M. Albrecht, and X.~X. Zhu, ``Ssl4eo-s12: A large-scale multimodal, multitemporal dataset for self-supervised learning in earth observation [software and data sets],'' \emph{IEEE Geoscience and Remote Sensing Magazine}, vol.~11, no.~3, pp. 98--106, 2023.

\bibitem{selfs}
Y.~Wang, C.~M. Albrecht, N.~A.~A. Braham, L.~Mou, and X.~X. Zhu, ``Self-supervised learning in remote sensing: A review,'' \emph{IEEE Geoscience and Remote Sensing Magazine}, vol.~10, no.~4, pp. 213--247, 2022.

\bibitem{1df9a7c2964043d5b80a06fcde793bbd}
S.~{De Saeger}, R.~Guelinckx, P.~Oosterlynck, A.~{De Bruyn}, K.~Debusschere, P.~Dhaluin, R.~Erens, P.~Hendrickx, D.~Hennebel, I.~Jacobs, M.~Kumpen, J.~Opdebeeck, T.~Spanhove, W.~Tamsyn, F.~{Van Oost}, G.~{Van Dam}, M.~{Van Hove}, C.~Wils, and D.~Paelinckx, \emph{\BIBforeignlanguage{Nederlands}{Biologische Waarderingskaart en Natura 2000 Habitatkaart, uitgave 2020}}, ser. Rapporten van het Instituut voor Natuur- en Bosonderzoek.\hskip 1em plus 0.5em minus 0.4em\relax Belgi{\"e}: Instituut voor Natuur- en Bosonderzoek, 2020, no.~35.

\bibitem{henighan2020scaling}
T.~Henighan, J.~Kaplan, M.~Katz, M.~Chen, C.~Hesse, J.~Jackson, H.~Jun, T.~B. Brown, P.~Dhariwal, S.~Gray \emph{et~al.}, ``Scaling laws for autoregressive generative modeling,'' \emph{arXiv preprint arXiv:2010.14701}, 2020.

\bibitem{hernandez2021scaling}
D.~Hernandez, J.~Kaplan, T.~Henighan, and S.~McCandlish, ``Scaling laws for transfer,'' \emph{arXiv preprint arXiv:2102.01293}, 2021.

\bibitem{tan2019efficientnet}
\BIBentryALTinterwordspacing
M.~Tan and Q.~Le, ``{E}fficient{N}et: Rethinking model scaling for convolutional neural networks,'' in \emph{Proceedings of the 36th International Conference on Machine Learning}, ser. Proceedings of Machine Learning Research, K.~Chaudhuri and R.~Salakhutdinov, Eds., vol.~97.\hskip 1em plus 0.5em minus 0.4em\relax PMLR, 09--15 Jun 2019, pp. 6105--6114. [Online]. Available: \url{https://proceedings.mlr.press/v97/tan19a.html}
\BIBentrySTDinterwordspacing

\bibitem{rosenfeld2021predictability}
J.~S. Rosenfeld, J.~Frankle, M.~Carbin, and N.~Shavit, ``On the predictability of pruning across scales,'' in \emph{International Conference on Machine Learning}.\hskip 1em plus 0.5em minus 0.4em\relax PMLR, 2021, pp. 9075--9083.

\bibitem{zhang2022resnest}
H.~Zhang, C.~Wu, Z.~Zhang, Y.~Zhu, H.~Lin, Z.~Zhang, Y.~Sun, T.~He, J.~Mueller, R.~Manmatha \emph{et~al.}, ``Resnest: Split-attention networks,'' in \emph{Proceedings of the IEEE/CVF conference on computer vision and pattern recognition}, 2022, pp. 2736--2746.

\bibitem{zhai2022scaling}
X.~Zhai, A.~Kolesnikov, N.~Houlsby, and L.~Beyer, ``Scaling vision transformers,'' in \emph{Proceedings of the IEEE/CVF conference on computer vision and pattern recognition}, 2022, pp. 12\,104--12\,113.

\bibitem{zhou2018unet++}
Z.~Zhou, M.~M. Rahman~Siddiquee, N.~Tajbakhsh, and J.~Liang, ``Unet++: A nested u-net architecture for medical image segmentation,'' in \emph{Deep Learning in Medical Image Analysis and Multimodal Learning for Clinical Decision Support: 4th International Workshop, DLMIA 2018, and 8th International Workshop, ML-CDS 2018, Held in Conjunction with MICCAI 2018, Granada, Spain, September 20, 2018, Proceedings 4}.\hskip 1em plus 0.5em minus 0.4em\relax Springer, 2018, pp. 3--11.

\bibitem{he2016deep}
K.~He, X.~Zhang, S.~Ren, and J.~Sun, ``Deep residual learning for image recognition,'' in \emph{Proceedings of the IEEE conference on computer vision and pattern recognition}, 2016, pp. 770--778.

\bibitem{NEURIPS2020_1457c0d6}
\BIBentryALTinterwordspacing
T.~Brown, B.~Mann, N.~Ryder, M.~Subbiah, J.~D. Kaplan, P.~Dhariwal, A.~Neelakantan, P.~Shyam, G.~Sastry, A.~Askell, S.~Agarwal, A.~Herbert-Voss, G.~Krueger, T.~Henighan, R.~Child, A.~Ramesh, D.~Ziegler, J.~Wu, C.~Winter, C.~Hesse, M.~Chen, E.~Sigler, M.~Litwin, S.~Gray, B.~Chess, J.~Clark, C.~Berner, S.~McCandlish, A.~Radford, I.~Sutskever, and D.~Amodei, ``Language models are few-shot learners,'' in \emph{Advances in Neural Information Processing Systems}, H.~Larochelle, M.~Ranzato, R.~Hadsell, M.~Balcan, and H.~Lin, Eds., vol.~33.\hskip 1em plus 0.5em minus 0.4em\relax Curran Associates, Inc., 2020, pp. 1877--1901. [Online]. Available: \url{https://proceedings.neurips.cc/paper_files/paper/2020/file/1457c0d6bfcb4967418bfb8ac142f64a-Paper.pdf}
\BIBentrySTDinterwordspacing

\bibitem{chen2014deep}
Y.~Chen, Z.~Lin, X.~Zhao, G.~Wang, and Y.~Gu, ``Deep learning-based classification of hyperspectral data,'' \emph{IEEE Journal of Selected topics in applied earth observations and remote sensing}, vol.~7, no.~6, pp. 2094--2107, 2014.

\bibitem{li2016hyperspectral}
W.~Li, G.~Wu, F.~Zhang, and Q.~Du, ``Hyperspectral image classification using deep pixel-pair features,'' \emph{IEEE Transactions on Geoscience and Remote Sensing}, vol.~55, no.~2, pp. 844--853, 2016.

\end{thebibliography}
\bibliographystyle{IEEEtran}

\newpage

\end{document}